%% file: main_jmlr.tex
\documentclass[twoside,11pt]{article}

\usepackage{jmlr2e}
\usepackage{amsmath,bm,float}
\usepackage{multirow}
\usepackage{todonotes}
\usepackage{subcaption}
\usepackage{array}
\PassOptionsToPackage{hyphens}{url}\usepackage{hyperref}

\newcommand{\mbf}[1]{\mathbf{#1}}
\newcolumntype{P}[1]{>{\raggedright\arraybackslash}p{#1}}
\newcommand{\cut}[1]{}

\DeclareMathSymbol{:}{\mathord}{operators}{"3A}

\newcommand{\jmlrtitle}[0]{A Survey on Principles, Models and Methods for Learning from Irregularly Sampled Time Series}

\ShortHeadings{Principles, Models and Methods for Learning from Irregularly Sampled Time Series}{Shukla and Marlin}
\firstpageno{1}

\begin{document}

\title{\jmlrtitle}
\author{\name Satya Narayan Shukla \email snshukla@cs.umass.edu \\
       \addr College of Information and Computer Sciences\\
       University of Massachusetts Amherst \\
       Amherst, MA 01003, USA
       \AND
       \name Benjamin M. Marlin \email marlin@cs.umass.edu \\
       \addr College of Information and Computer Sciences\\
       University of Massachusetts Amherst \\
       Amherst, MA 01003, USA
}


\maketitle

\begin{abstract}
Irregularly sampled time series data arise naturally in many application domains including biology, ecology, climate science, astronomy, and health. Such data represent fundamental challenges to many classical models from machine learning and statistics due to the presence of non-uniform intervals between observations. However, there has been significant progress within the machine learning community over the last decade on developing specialized models and architectures for learning from irregularly sampled univariate and multivariate time series data. In this survey, we first describe several axes along which approaches to learning from irregularly sampled time series differ including what data representations they are based on, what modeling primitives they leverage to deal with the fundamental problem of irregular sampling, and what inference tasks they are designed to perform. We then survey the recent literature organized primarily along the axis of modeling primitives. We describe approaches based on temporal discretization, interpolation, recurrence, attention and structural invariance. We discuss similarities and differences between approaches and highlight primary strengths and weaknesses.
\end{abstract}


\input{intro}

\input{representation}
\input{tasks}
\input{modeling_primitives}

\input{discretization}
\input{interpolation}

\input{recurrence}
\input{attention}
\input{set}

\input{performance}
\input{conclusions}


\vskip 0.2in
\bibliography{references}

\appendix
\input{dataset}

\input{appendix}

\end{document}

%% file: intro.tex
\section{Introduction}
An irregularly sampled time series is a sequence of time-value pairs with non-uniform intervals between successive time points. This paper surveys approaches for modeling and learning from irregularly sampled time series generated from underlying continuously varying, real-valued univariate or multivariate functions.  
Such time series data naturally occur in domains where the observation process is constrained to a degree that prohibits regular observation of continuously varying phenomena. 

Irregularly sampled time series occur in a number of scientific and industrial domains and is also a prominent feature of some types of data in the health domain. 
For example, in astronomy, measurements of properties such as the spectra of celestial objects are taken at times determined by seasonal and weather conditions as well as the availability of time on required instruments \citep{scargle-astro1982}. 
In electronic health records data, vital signs and other measurements are recorded at time points determined by factors that may include the type of measurement, the patient's state of health, and the availability of clinical staff  \citep{marlin-ihi2012}. 
Irregularly sampled time series data commonly occur in a range of other areas with similarly complex observation processes including in biology \citep{ruf1999lomb}, climate science \citep{schulz1997spectrum}, geology \citep{kira2011}, finance \citep{manimaran2006}, economics \citep{kling1985}, meteorology \citep{mudelsee2002}, and traffic analysis \citep{qing2010}.

\begin{figure}[t]
    \centering
    \includegraphics[width=\textwidth]{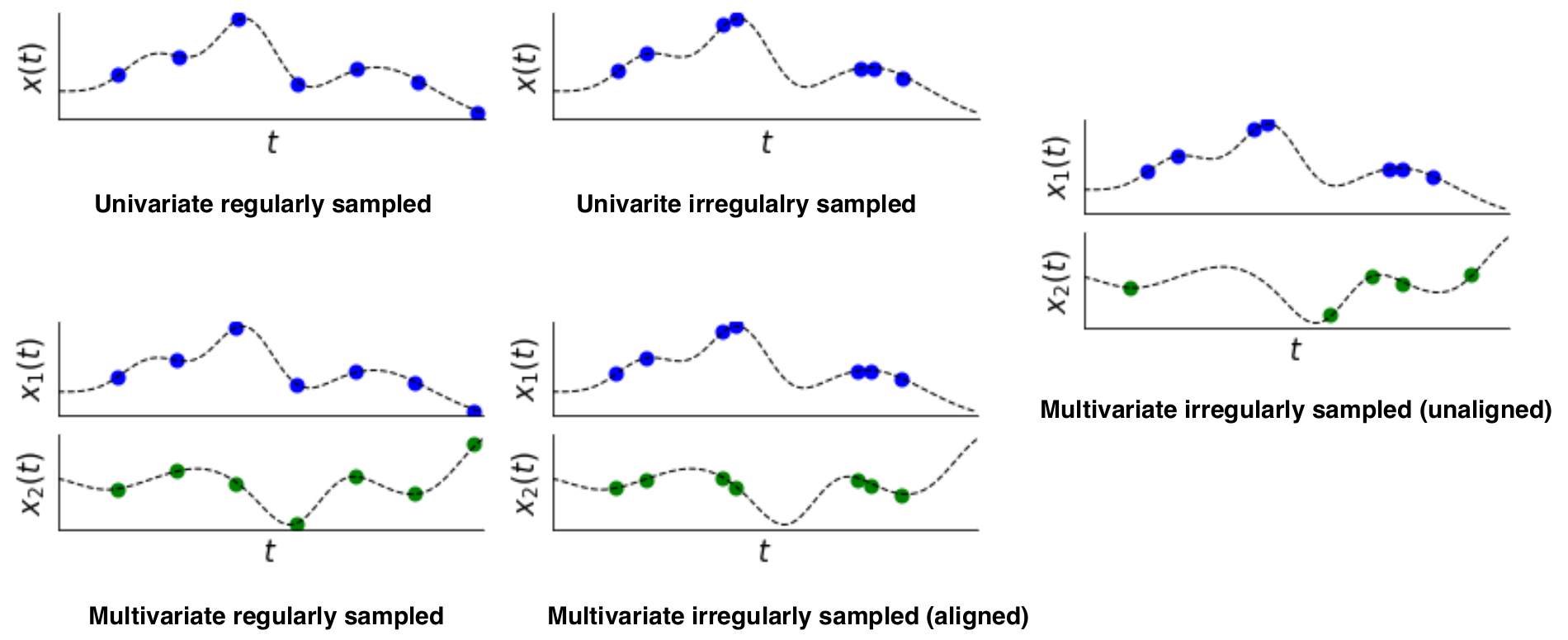}
    \caption{Illustration of regularly and irregularly sampled univariate and two-dimensional multivariate  time series. Univariate irregularly sampled time series data are characterized by variable time intervals between observations. Multivariate irregularly sampled time series are also characterized by variable time intervals between observations, but can have aligned or unaligned observation times across dimensions. When observation time points are unaligned across dimensions, different dimensions also often contain different numbers of observations. Finally, while this figure only illustrates  single univariate or multivariate time series, it is important to note that in data sets containing multiple univariate or multivariate time series as data cases, the observation times in different data cases are also typically unaligned and different data cases often also have different numbers of observations.}
    \label{fig:ists}
\end{figure}

Irregularly sampled time series data (as shown in Figure \ref{fig:ists}) present fundamental challenges to many classical models from machine learning and statistics. To illustrate these challenges, consider the case of a supervised learning task where a model takes as input an irregularly sampled time series and must predict a scalar output. To learn such a model, we assume access to a data set $\mathcal{D}$ where each instance is a tuple $(\mbf{s}_n,y_n)$. Here $\mbf{s}_n$ represents the irregularly sampled time series and $y_n$ represents the corresponding prediction target. The primary challenges in modeling such data are:



\begin{itemize}
    \item[(i)] Irregular sampling  - the presence of variable gaps between successive observation time points within a time series.
    \item[(ii)] Variable number of observations  - the total number of observations across dimensions can be different for different data cases. 
    \item[(iii)] Lack of alignment  - different dimensions of a single multivariate time series can be observed at a different collection of time points. The collection of observation times across dimensions can also differ between data cases. 
\end{itemize}

These features of irregularly sampled time series data invalidate the assumption of a coherent fixed-dimensional feature space, which underlies most basic supervised and unsupervised learning models including K-nearest neighbours \citep{knn}, decision trees \citep{decisiontrees}, linear  and logistic regression \citep{hastie01statisticallearning, hosmer2013applied}, linear, polynomial and RBF-kernel support vector machines \citep{cortes1995}, multi-layer perceptrons \citep{hastie01statisticallearning}, K-means \citep{kmeans}, mixtures models (with standard component distributions) \citep{mixture}, factor analysis \citep{factoranalysis}, PCA \citep{pca}, autoencoders \citep{autoencoder}, and more.  However, there has been significant progress over the last decade on developing specialized models and architectures for learning from irregularly sampled multivariate time series data within the machine learning community. 

This paper presents a survey of many of these approaches that focuses on categorizing methods in terms of three key properties including the underlying data representation they use, the fundamental mechanism they use to accommodate irregular sampling, and the types of machine learning tasks to which they can be applied. 
We identify three primary underlying data representations for irregularly sampled time series that we refer to as \textit{series-based}, \textit{vector-based}, and  
\textit{set-based} representations. In the series-based representation, a multivariate irregularly sampled time series is viewed as consisting of a collection of univariate time series, each with its own collection of observation times and values. In the vector-based representation, a multivariate time series is viewed as a time series of vector-valued observations. When observations for different dimensions are not temporally aligned, the result is missing values in the vector-valued observations. The set-based representation views a multivariate irregularly sampled time series as a set of  tuples of the form $(t,d,x)$ where $t$ is a time point, $d$ is a dimension,  and $x$ is the observed value of dimension $d$ at time $t$. As we will see, different approaches are based on different of these representations, leading to methods with different capabilities and limitations.

\begin{figure}[t]
    \centering
    \includegraphics[width=1\textwidth]{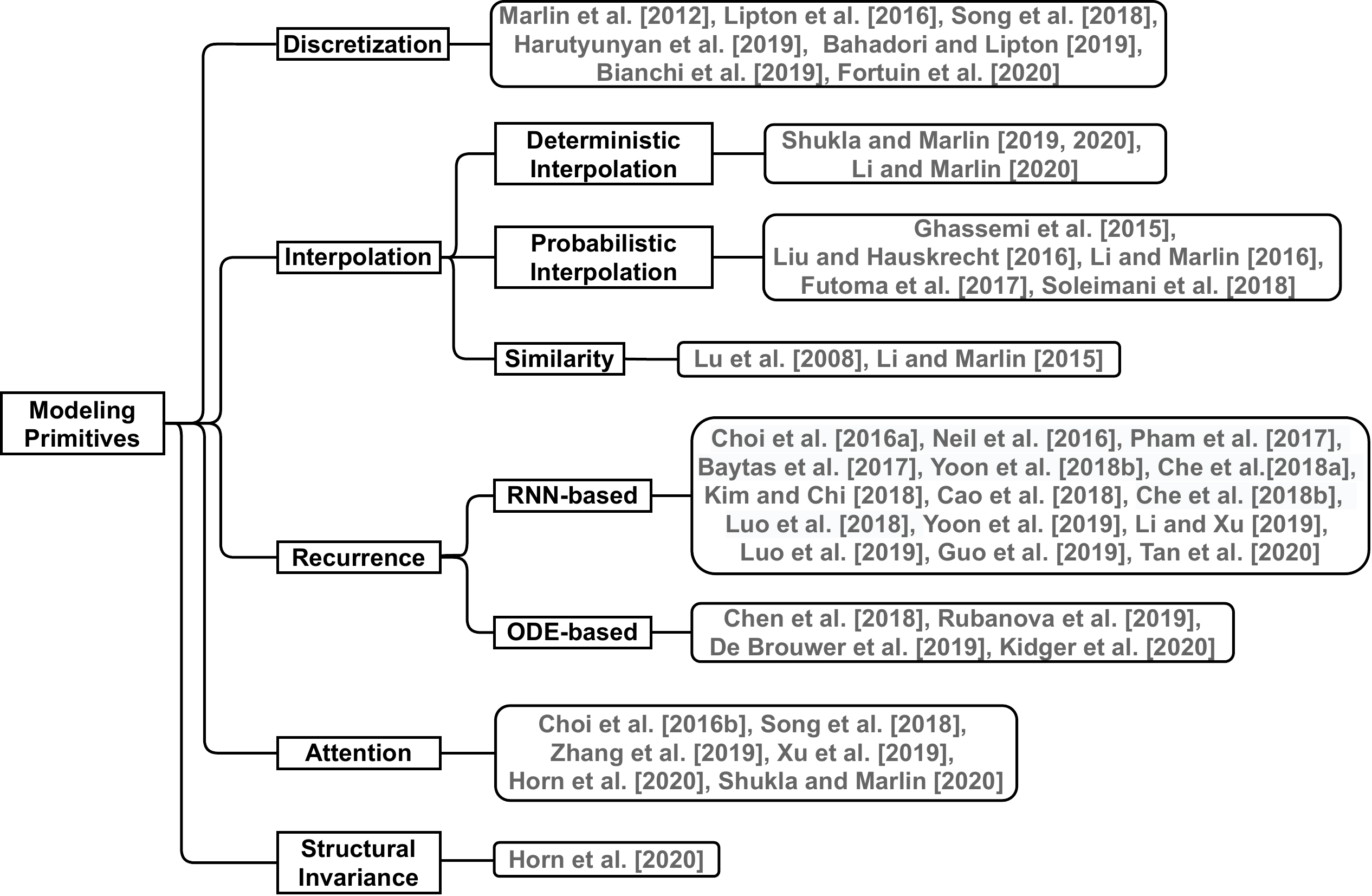}
    \caption{This figure illustrates a taxonomy of methods based on modeling primitives. We define modeling primitives to be the fundamental approach that methods use to accommodate irregular sampling. We identify five high-level modeling primitives including discretization, interpolation, recurrence, attention and structural invariance.  }
    \label{fig:categorization}
\end{figure}

Different approaches also leverage different ``modeling primitives" for accommodating irregular sampling. We define modeling primitives to be the basic building blocks or modules leveraged in larger and more complex models. We identify several such modeling primitives that specifically provide the interface between irregularly sampled time series data and more standard model components. These modeling primitives fall into several categories including approaches based on temporal discretization, interpolation, similarity, recurrence, attention and structural invariance. We categorize approaches in terms of their use of such modeling primitives, which can sometimes be obscured in larger and more complex models. In Figure \ref{fig:categorization}, we provide a categorization of different approaches under a taxonomy based on modeling primitives. 

The third categorization that we focus on is the set of inference tasks that a given approach is designed to solve. These tasks include detection, prediction, filtering, smoothing, interpolation, and forecasting. Some approaches can be applied to multiple of these tasks, while others can not. Understanding the range of problems a given approach can be applied to in a valid way is obviously important, but can again be unclear for some larger and more complex models.    

%


The rest of this survey is organized as follows. In Section \ref{sec:representation}, we describe the categorization of representations for irregularly sampled time series. In Section \ref{tasks}, we define modeling tasks for irregularly sampled time series. In Section \ref{modeling_primitives}, we define modeling primitives for irregularly sampled time series.  In Sections \ref{sec:discretization_approaches} to \ref{sec:invariance_approaches}, we present a detailed discussion of specific models and approaches organized by their primary modeling primitives including temporal discretization, interpolation, similarity, recurrence, attention and structural invariance. In Section \ref{sec:performance}, we discuss the predictive performance of methods. Finally, in Appendix \ref{sec:datasets} we describe data sets commonly used to evaluate models for irregularly sampled time series while in Appendix \ref{sec:code}, we provide links to open source implementations of many of the approaches described in this survey.

%% file: representation.tex
\section{Data Representations for Irregularly Sampled Time Series}
\label{sec:representation}

There are several possible data representations for multivariate irregularly sampled time series. While these representations are equivalent, they expose different properties and suggest different approaches to modeling. In all cases we will assume that a data set contains $N$ data cases and that each data case consists of observations of $D$ different variables through time.  We begin by describing what we will refer to as the series-based representation followed by the vector-based representation and the set-based representation.

\subsection{Series-Based Representation}
\label{sec:series}
\begin{figure}[t]
    \centering
    \includegraphics[width=0.75\textwidth]{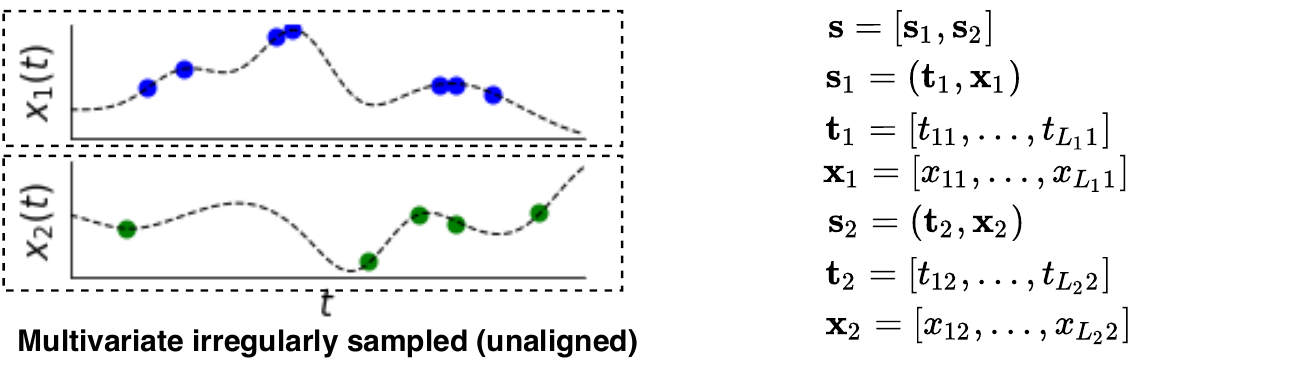}
    \caption{Illustration of the series-based representation for a two-dimensional multivariate irregularly sampled time series. In the series-based representation, a $d$-dimensional multivariate irregularly sampled time series $\mbf{s}=[\mbf{s}_1,...,\mbf{s}_D]$ is represented as a collection of univariate irregularly sampled time series, one per dimension. Here $\mbf{s}_d=(\mbf{t}_d,\mbf{x}_d )$ indicates the time series for dimension $d$. $\mbf{t}_d$ indicates the collection of time points with observed values for dimension $d$ while $\mbf{x}_d$ indicates the corresponding collection of observed values. 
    }
    \label{fig:series}
\end{figure}

As shown in Figure \ref{fig:series}, the series-based representation of a multivariate irregularly sampled time series data set views the data for a single data case $n$ as a collection of $D$ univariate irregularly sampled time series with one item in the collection for each of the $D$ dimensions. We define $L_{dn}$ to be the number of observations of variable $d$ for data case $n$. We define $\mbf{t}_{dn}=[t_{1dn},...,t_{L_{dn}dn}]$ to be the collection of time points at which variable $d$ is observed for data case $n$. We define $\mbf{x}_{dn}=[x_{1dn},...,x_{L_{dn}dn}]$ to be the corresponding collection of observed values.\footnote{ We note that we require that the observations are indexed in time order i.e., $t_{idn} < t_{jdn}$ for $i < j$.} The data for time series $n$ and variable $d$ is then a univariate irregularly sampled time series $\mbf{s}_{dn}=(\mbf{t}_{dn}, \mbf{x}_{dn})$. We define $\mbf{s}_n = [\mbf{s}_{1n},...,\mbf{s}_{Dn}]$ to be the complete multivariate irregularly sampled time series for data case $n$. 

We note that in this representation, there is no missing data. We represent only the observations available for each dimension. Further, this structure exposes the fact that in the fully general case of multivariate irregular sampling, different dimensions can be observed at different collections of time points with different total numbers of observations for the same data case. We also note that this representation can be wasteful in scenarios where all variables are observed at each time point. This motivates the vector-based representation, which we introduce next.

\subsection{Vector-Based Representation}
\label{sec:vector}
\begin{figure}[t]
    \centering
    \includegraphics[width=0.9\textwidth]{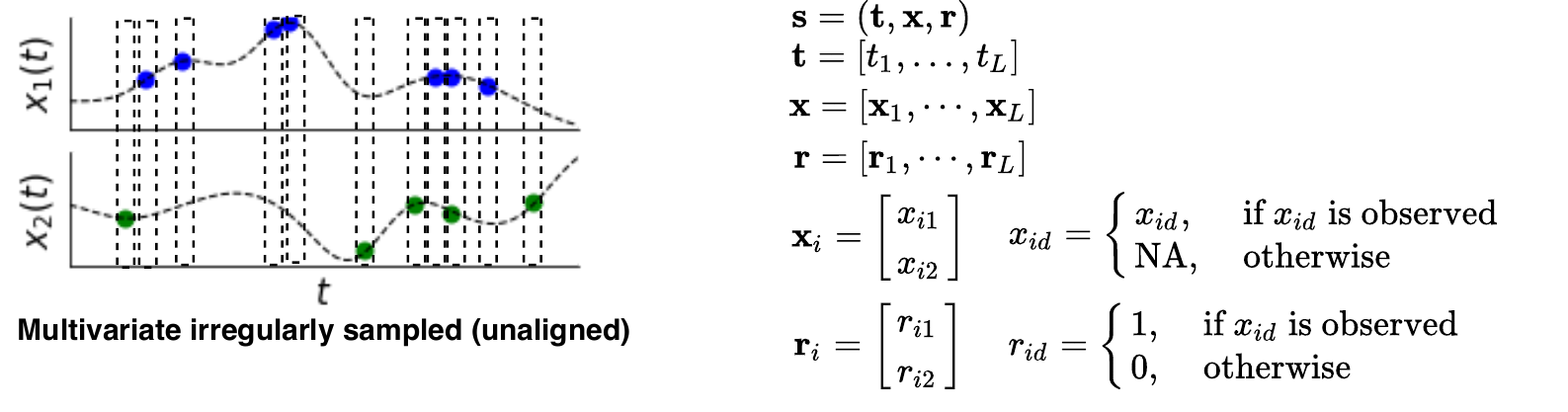}
    \caption{Illustration of the vector-based representation for multivariate irregularly sampled time series in the case of two dimensions with unaligned observation times. In this representation, there is a single collection of time points $\mbf{t}$. At each time point $t_i$, we define a $D$-dimensional vector-valued observation $\mbf{x}_i$. In the general case, not all dimensions of $\mbf{x}_i$ are observed, leading to the need to explicitly represent which dimensions are observed and which are missing. Following \citet{little2014statistical}, we introduce a $D$-dimensional binary response indicator vector $\mbf{r}_i$ at each time point $t_i$ to indicate which dimensions are observed and which are missing. The complete representation for the time series is thus $\mbf{s}=(\mbf{t},\mbf{x},\mbf{r})$ where $\mbf{t}$ is the collection of observation times, $\mbf{x}$ is the collection of vector-valued observations (with missing values), and $\mbf{r}$ is the collection of response indicators.
    }
    \label{fig:vector}
\end{figure}

As shown in Figure \ref{fig:vector}, the vector-based representation  of a multivariate irregularly sampled time series data set views the data for a single data case $n$ as a single series of $D$ dimensional vectors. We define $L_{n}$ to be the number of time points with observations for data case $n$. We define $\mbf{t}_{n}=[t_{1n},...,t_{L_{n}n}]$ to be the collection of time points with observations for data case $n$. As with the series based representation, we assume observations are indexed in time order.
Next, we define $\mbf{x}_{n}=[\mbf{x}_{1n},...,\mbf{x}_{L_{n}n}]$ to be the corresponding collection of $D$-dimensional vector-valued observations. We again let $x_{idn}$ be the observed value of dimension $d$ at time point $t_{in}$. We define $\mbf{s}_n = (\mbf{t}_{n},\mbf{x}_{n})$ to be the irregularly sampled time series for data case $n$. Under this representation, different data cases $n$ can still have different collections of observation time points as well as different numbers of observations. However, the observations across dimensions for a single data case are assumed to be aligned in time. Compared to the series-based representation, the vector-based representation is thus substantially more space efficient when the data are fully observed across dimensions at each time point.

When the vector $\mbf{x}_{in}$ is not fully observed at time $t_{in}$, the vector-based representation results in the need to represent missing data. Following \citet{little2014statistical}, the standard approach to representing which elements of $\mbf{x}_n$ are observed and which are missing is to introduce an auxiliary response indicator series $\mbf{r}_n=[\mbf{r}_{1n},...,\mbf{r}_{L_{n}n}]$ where ${r}_{idn}=1$ if the value of  ${x}_{idn}$ is observed and ${r}_{idn}=0$ otherwise. The full vector-based representation of an irregularly sampled time series with incomplete observations is thus  $\mbf{s}_n = (\mbf{t}_{n},\mbf{x}_{n}, \mbf{r}_{n})$. 

\subsection{Set-Based Representation}
\begin{figure}[t]
    \centering
    \includegraphics[width=0.9\textwidth]{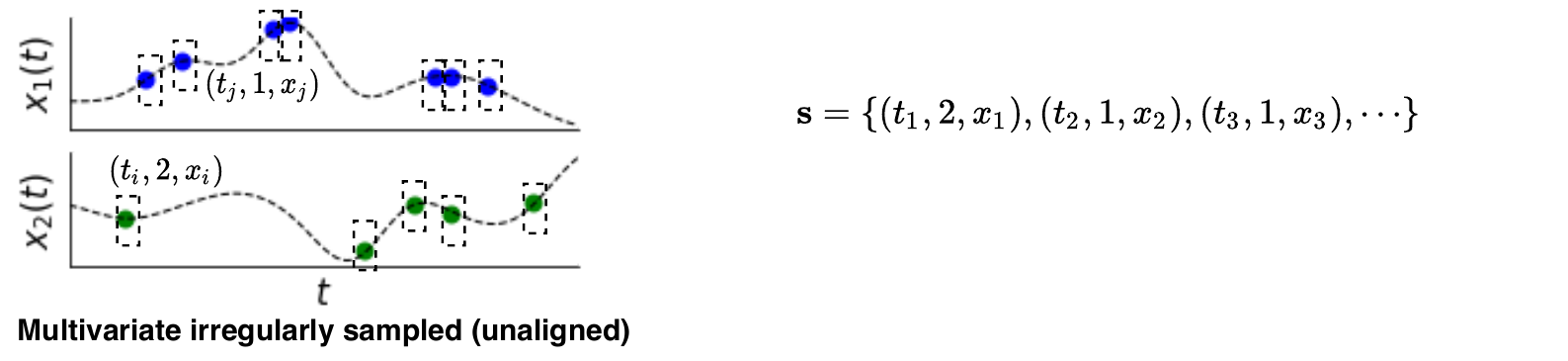}
    \caption{Illustration of the set-based representation for multivariate irregularly sampled time series in the case of two dimensions. In this representation, a $D$-dimensional multivariate irregularly sampled time series is represented as a set of (time, dimension, value) tuples, one for each observation.}
    \label{fig:set}
\end{figure}

As shown in Figure \ref{fig:set}, the set-based representation of a multivariate irregularly sampled time series data set views the data for a single data case $n$ as set of time-dimension-value tuples of the form $(t_{in},d_{in},x_{in})$. Here we let $L_n$ represent the total number of observations across all time points and dimensions. The irregularly sampled time series for data case $n$ is thus represented as $\mbf{s}_n = \{(t_{in},d_{in},x_{in})| 1\leq i\leq L_n\}$. Like the series-based representation, the set-based representation does not require explicit representation of missing data. Unlike both the the series and vector-based representation, the time ordering of the data is not explicitly reflected in the structure of the set-based representation. 

%% file: tasks.tex
\section{Inference Tasks}
\label{tasks}

\begin{figure}[t]

\begin{subfigure}[b]{0.325\textwidth}
\raggedleft
   \includegraphics[width=1\textwidth, bb = 10 25 160 180, clip]{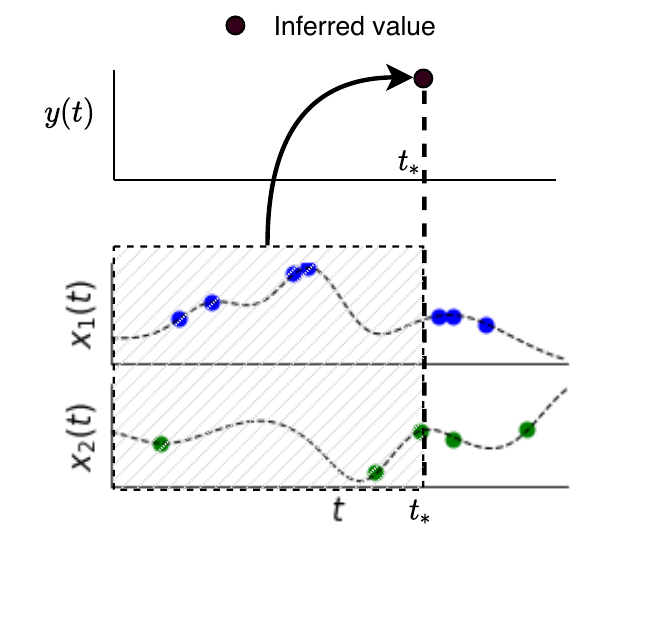}
   \caption{Detection}
\end{subfigure}
\begin{subfigure}[b]{0.325\textwidth}
\raggedleft
   \includegraphics[width=1\textwidth, bb = 10 20 170 180, clip]{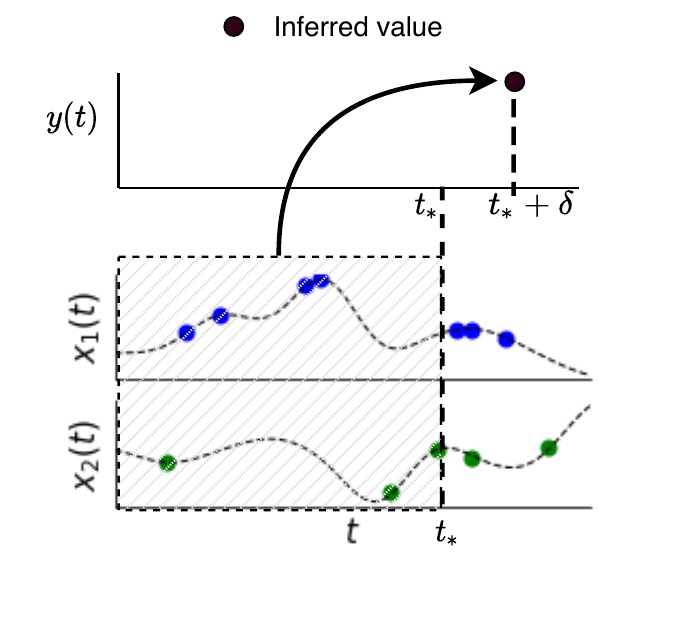}
   \caption{Prediction}
\end{subfigure}
\begin{subfigure}[b]{0.325\textwidth}
\raggedright
   \includegraphics[width=1.2\textwidth, bb = 5 20 170 170, clip]{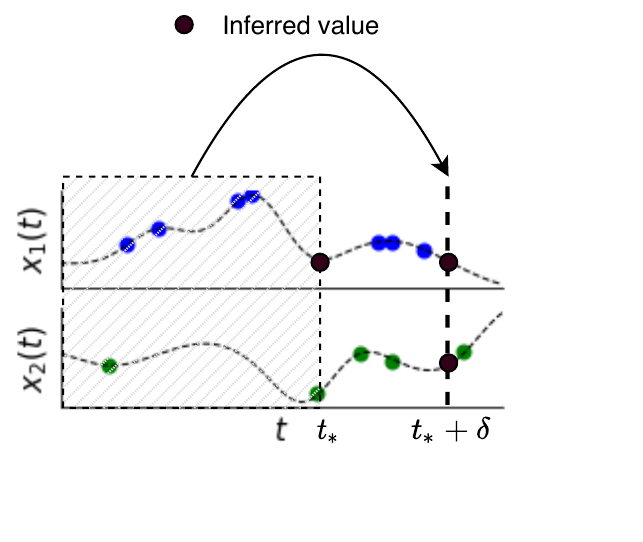}
   \caption{Forecasting}
\end{subfigure}

\begin{subfigure}[b]{0.325\textwidth}
  \includegraphics[width=1\textwidth, bb = 5 30 140 150, clip]{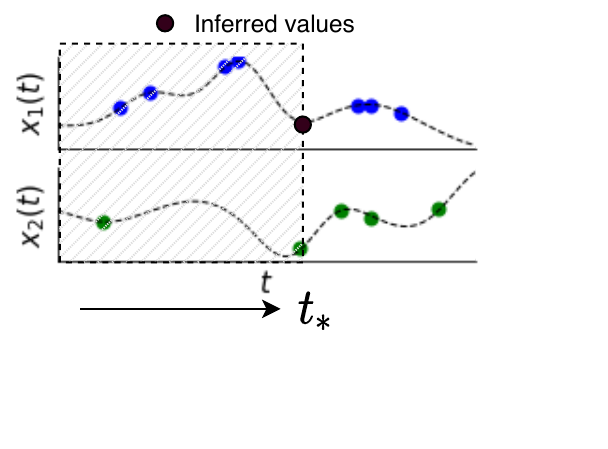}
   \caption{Filtering}
\end{subfigure}
\begin{subfigure}[b]{0.325\textwidth}
  \includegraphics[width=1\textwidth, bb = 5 30 140 150, clip]{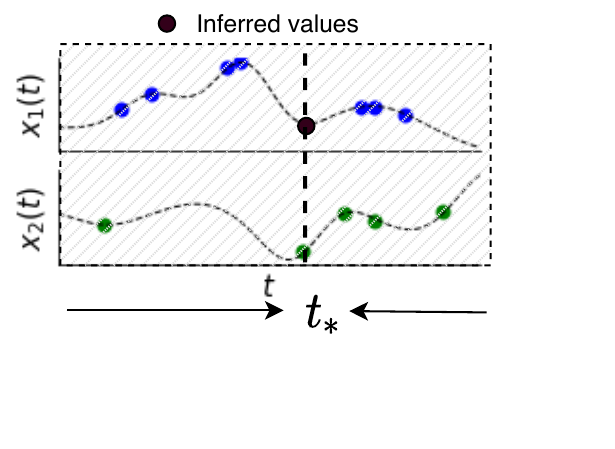}
  \caption{Smoothing}
\end{subfigure}
\begin{subfigure}[b]{0.325\textwidth}
  \includegraphics[width=1\textwidth, bb = 5 30 140 150, clip]{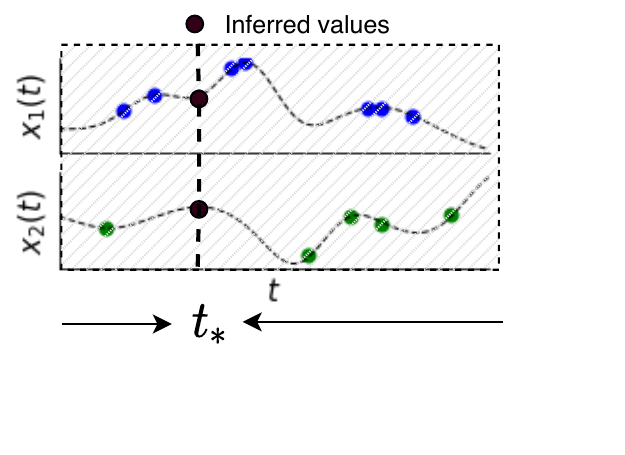}
  \caption{Interpolation}
\end{subfigure}

\caption{This figure illustrates time series inference tasks. (a) The detection task involves predicting the target values $y[t_*]$ at time $t_*$ conditioning on the observations available up to and including time $t_*$. (b) The prediction task requires inferring the prediction target value $y[t_* + \delta]$ at time $t_* + \delta$ for $\delta>0$ by conditioning on the observations available up to and including time $t_*$. 
(c) The forecasting task requires inferring $\mbf{x}[t_* + \delta]$ by conditioning on the observations
up to an including time $t_*$.
(d) The filtering task requires inferring missing variables at time $t_*$ by conditioning on the observations up to an including time $t_*$. (e) The smoothing task requires inferring missing variables at time $t_*$ using all observed data. (f) The interpolation task requires inferring the values of $x[t_*]$ at time $t_*$ using all observed data.}

\label{fig:tasks}
\end{figure}

In this section, we define time series inference tasks. While these tasks are not specific to the case of irregularly sampled time series, carefully specifying inference tasks is necessary to properly categorize models for learning from irregularly sampled time series in terms of the tasks that they can perform. In the definition of these tasks, it will be important to specify the time ranges that are conditioned on for a given input irregularly sampled time series. We will use the notation $\mbf{s}[:t]$ to denote all of the data contained in time series $\mbf{s}$ that are observed up to and including time $t$. We will use $\mbf{x}[t]$ to refer to the vector of observations available at time $t$, some (or all) of which may be missing. For brevity, we will use $\mbf{x}^m[t]$ to indicate the sub-vector of $\mbf{x}[t]$ whose values are missing at time $t$. Some of the tasks we consider are supervised in the sense that they involve learning to infer values that are distinct from the input irregularly sampled time series. We denote these outputs at time $t$ by $\mbf{y}[t]$. Below we define and discuss the six inference tasks shown in Figure \ref{fig:tasks}. 


\begin{definition}
{\bf Detection}: Inferring prediction target values $\mbf{y}[t_*]$ at time $t_*$ conditioning on the observations $\mbf{s}[:t_*]$ available up to and including time $t_*$.  
\end{definition}

The detection task is to infer the value of the prediction target variable at time $t_*$. All time series data observed up to and including time $t_*$ can be conditioned on when making this inference. We note that in machine learning, the inference for any quantity that is not known is often referred to as a ``prediction", but in this context we reserve the term ``prediction" to refer to a task where the inference is for the value of the output variable at a time that is in the relative future of the time point $t_*$ at which the inference is made, as we describe next.

\begin{definition}
{\bf Prediction}: Inferring prediction target values $\mbf{y}[t_*+\delta]$ 
at time $t_* + \delta$ (for $\delta>0$) conditioning on the observations $\mbf{s}[:t_*]$ available up to and including time $t_*$.  
\end{definition}

In the prediction problem, $t_*$ refers to the time point at which the prediction is made and $t_*+\delta$ refers to the time point at which the value of the target variable is to be predicted. We assume $\delta>0$. The case  $\delta=0$ recovers the detection problem.

We note that in some supervised problems, the prediction target variables may not have time stamps explicitly associated with them. Such problems include whole time series regression and classification where an input time series $\mbf{s}$ is associated with a single scalar output $y$. These problems are equivalent in structure to the prediction problem defined above where all available data are allowed to be conditioned on when inferring the target value.

\begin{definition}
{\bf Filtering}: Inferring missing variables $\mbf{x}^m[t_*]$ at time $t_*$ by conditioning on the observations  $\mbf{s}[:t_*]$ up to an including time $t_*$.
\end{definition}

The filtering task is the analog of the detection task but where the inference of interest is about the values of the multivariate time series itself. In the irregularly sampled setting, it may be that some dimensions of $\mbf{x}[t_*]$ are observed at time $t_*$. In  this case, these values and any values observed before this time point can be used to infer the unobserved values $\mbf{x}^m[t_*]$. 

\begin{definition}
{\bf Smoothing}: Inferring the values of $\mbf{x}^m[t_*]$ at time $t_*$ using the observed data in $\mbf{s}$. 
\end{definition}

The smoothing task is similar to the filtering task except that all observed data can be used to infer the unobserved values $\mbf{x}^m[t_*]$. This includes observations in both the relative past and future of $\mbf{x}^m[t_*]$. 

\begin{definition}
{\bf Interpolation}: Inferring the values of $\mbf{x}[t_*]$ at time $t_*$ using the observed data in $\mbf{s}$. 
\end{definition}

The interpolation task is similar to the filtering task except that in the  interpolation task, we infer the values of all variables at time $t_*$, not just those that are unobserved. This distinction is only relevant in scenarios where the time point $t_*$ coincides exactly with an observation time point contained in the time series $\mbf{s}$. 

\begin{definition}
{\bf Forecasting}: Inferring $\mbf{x}[t_*+\delta]$ (for $\delta>0$) by conditioning on the observations  $\mbf{s}[:t_*]$ up to an including time $t_*$. 
\end{definition}

The forecasting task is the analog of the prediction task, but where we seek to predict the value of the input time series itself at a future time point. In this problem, $t_*$ again refers to the time point at which the forecast is made and $t_*+\delta$ for $\delta>0$ refers to the time point about which the forecast is made. All observations up to and including time $t_*$ can be used to compute a  forecast at time $t_*$. $\delta$ is often referred to as the forecast horizon, the amount of time into the future we are forecasting the value of the time series. 

%% file: modeling_primitives.tex
\section{Modeling Primitives for Irregularly Sampled Time Series}
\label{modeling_primitives}

While there has been a significant volume of work in recent years on modeling sparse and irregularly sampled time series data in the context of different tasks and applications, the number of fundamental approaches for accommodating irregular sampling itself is much more limited. In this section, we describe several ``modeling primitives" for accommodating irregular sampling. These modeling primitives are the basic building blocks or modules for dealing with irregular sampling that are leveraged in larger and more complex models and methods for solving specific tasks using irregularly sampled time series. We discuss a number of modeling primitives including discretization, interpolation, similarity, recurrence, attention and structural invariance. Figure \ref{fig:categorization} provides a taxonomy of modeling primitives and associated methods.

\subsection{Discretization}
\label{sec:discretization}
\begin{figure}[t]
 \centering
    \includegraphics[width=0.8\textwidth]{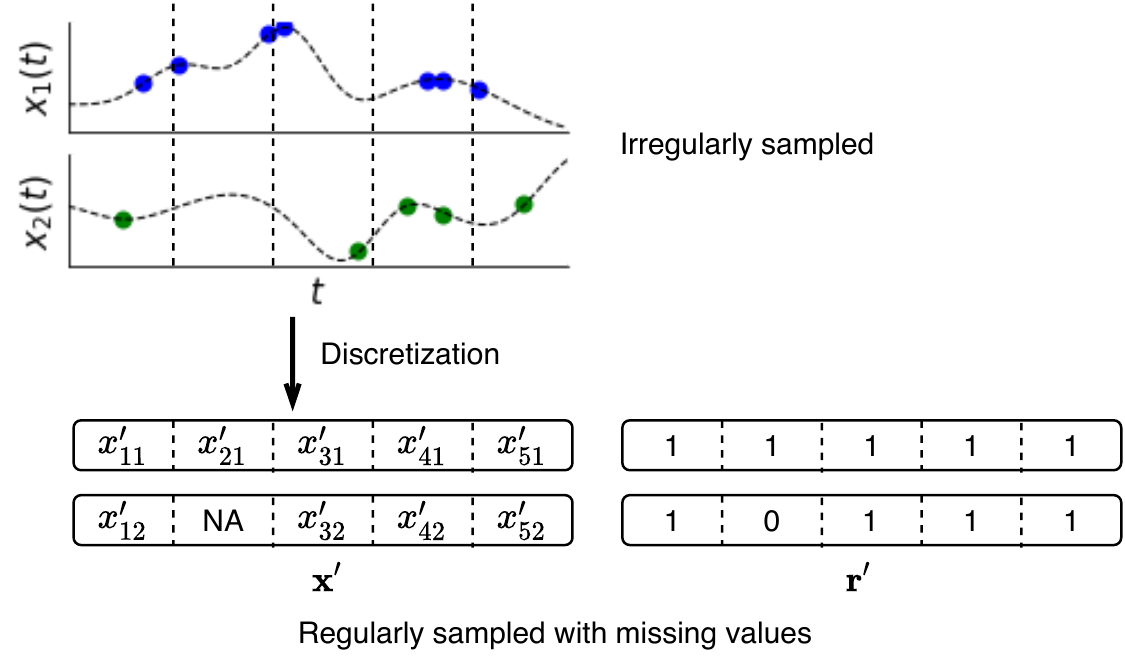}
    \caption{This figure illustrates the discretization of irregularly sampled time series into regularly spaced time series with missing values. The discretization process requires dividing the time axis into equal sized non-overlapping intervals and defining a value within each time interval based on the observed values falling within that interval. In cases where a given interval contains no observations on a particular dimension, the result is missing data. We can again represent these missing values via auxiliary response indicator vectors.}
    \label{fig:discretization}
\end{figure}
Temporal discretization is a basic modeling primitive used to convert irregularly sampled time series data into a regularly sampled time series, as shown in Figure \ref{fig:discretization}. To apply this primitive, one first defines a sequence of $K+1$ regularly spaced reference time points $\tau_0,...,\tau_{K}$. Given an irregularly sampled time series $\mbf{s}_n$, we define a new regularly sampled time series in the vector-based representation with missing data indicators $\mbf{s}'_n=(\mbf{t}'_n,\mbf{x}'_n,\mbf{r}'_{n})$. 
To begin, we let $\mbf{t}'_n=[t'_1,...,t'_K]$ for all $n$ where $t'_i = (\tau_{i-1}+\tau_{i})/2$.\footnote{We choose the midpoint of the interval $[\tau_{i-1},\tau_{i})$ to represent the interval, but the starting point or any other consistent choice could also be made.} 

Next, we define a function that maps the observed values of $\mbf{s}_n$ falling within the discretization window $[\tau_{i-1},\tau_i)$ to the vectors $\mbf{x}'_{in}$ and $\mbf{r}'_{in}$. As a first case, assume there are $m>0$ observations for dimension $d$ in $\mbf{s}_n$ within the interval $[\tau_{i-1},\tau_i)$ and let their indices be $\mbf{o}=[o_1,...,o_m]$. We require a function of the form  $\mbf{x}'_{idn} = f([\mbf{t}_{o_1n},...,\mbf{t}_{o_mn}],[\mbf{x}_{o_1dn},...,\mbf{x}_{o_mdn}])$ that maps the set of observations within the discretization window to a single representative value. A common approach is to set this function to be the average of the $m$ values. To complete the representation for this interval, we set $\mbf{r}'_{idn}=1$ to indicate that the interval has an observed value. 

In the second case where $m=0$, the original time series has no observations of dimension $d$ within the interval $[\tau_{i-1},\tau_i)$. In this case we set  $\mbf{r}'_{idn}=0$ to indicate no observations are available. The value set in $\mbf{x}'_{idn}$ is not consequential as any correct algorithm for this representation will not use values in  $\mbf{x}'_{idn}$ when $\mbf{r}'_{idn}=0$ \citep{little2014statistical}.  

As we can see, the discretization primitive provides a reduction from the problem of modeling irregularly sampled time series to the problem of modeling regularly sampled time series that may contain missing data. As the discretization windows become larger, the volume of missing data will generally decrease. However, more values will potentially also fall within the same discretization window leading to more aggregation. This may remove information needed to accurately perform some tasks. As discretization windows become shorter, aggregation effects are reduced, but the length of the time series increase as does the volume of missing data. Thus, the window size becomes an important hyperparameter of methods based on this approach. 

We note that when discretization results in significant volumes of  missing data, the problem of dealing with the resulting missing data can itself be highly non-trivial as the presence of incomplete data also violates the assumptions of most standard discriminative machine learning methods. Imputation methods can be used as a final stage of pre-processing to address the problem of missing data, but care is needed when reflecting uncertainty is important. A number of recent deep learning approaches have addressed the problem of providing flexible single and multiple imputation methods  \citep{gain,misgan,miwae,missing2018}. Many simpler methods are also commonly used for time series  including mean imputation, zero imputation and forward-filling.

\subsection{Interpolation}
\label{sec:interpolation}

Interpolation primitives provide an approach to accommodating irregular sampling that is closely related to discretization but provides improved flexibility with potentially fewer ad-hoc assumptions. Interpolation methods are often used to provide an interface between irregularly sampled time series data and models for either fixed-dimensional feature spaces or for variable length sequences. 

\begin{figure}[t]
    \centering
    \includegraphics[width=\textwidth]{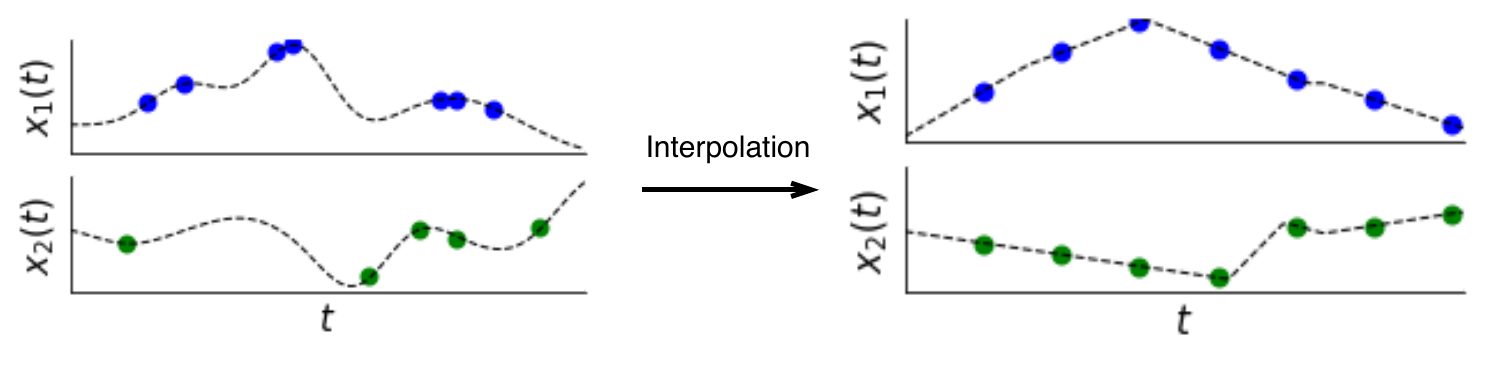}
    \caption{Illustration of interpolation in the case of a two-dimensional irregularly sampled time series. In this example, we illustrate the basic use of linear interpolation against a fixed set of reference time points.}
    \label{fig:interpolation}
\end{figure}
Similar to discretization, we begin by defining a set of $K$ reference time points $\bm{\tau}=[\tau_1,...,\tau_{K}]$. In the case of a length $L_n$ univariate irregularly sampled time series $\mbf{s}_n=(\mbf{t}_n,\mbf{x}_n)$, a basic kernel smoother \cite[Ch. 6]{hastie01statisticallearning} produces interpolated output $\mbf{x}'_{in}$ at reference time point $\tau_i$ as shown in Equation \ref{eq:interp}. The interpolated time series is then given by $\mbf{s}'_n=(\bm{\tau}, \mbf{x}'_n)$. The function $\kappa_{\theta}()$ is a similarity kernel that typically puts higher weight on pairs of time points $\tau_i, \mbf{t}_{jn}$ that are closer together in time. This function can depend on a set of learnable parameters $\theta$. The squared exponential kernel function $\kappa_{\theta}(t,t')=\alpha\exp(-\beta(t-t')^2)$ is a common choice in the literature for non-linear interpolation (with $\theta=[\alpha,\beta]$).
\begin{align}
\label{eq:interp}
    \mbf{x}'_{in} = \frac{\sum_j \kappa_{\theta}(\tau_i, \mbf{t}_{jn})\, \mbf{x}_{jn}}{\sum_j \kappa_{\theta}(\tau_i, \mbf{t}_{jn})}
\end{align}
For multivariate irregularly sampled time series, there are more possibilities in terms of the construction of interpolation methods as these methods can account for both correlation in time and correlation across different dimensions. One basic approach motivated by the series-based view of a multivariate time series is to separately interpolate each dimension of the time series using a univariate interpolation approach as shown in Equation \ref{eq:interp} and in Figure \ref{fig:interpolation}, but with the same set of reference time points used across all dimensions. This ignores cross-dimension correlations, but provides regularly spaced output to which further modeling can be applied to account for cross-dimensional correlations.  

One potential issue when inter-observation times exhibit significant variability is  variable uncertainty in interpolated values. The use of Gaussian process regression (GPR) models can provide a better alternative to deterministic interpolation methods in such cases \citep{gp2006}. The primary strength of GPR models is that they provide a joint posterior distribution over an arbitrary collection of reference time points $\bm{\tau}=[\tau_1,...,\tau_{K}]$ conditioned on an input irregularly sampled time series $\mbf{s}_n$. This posterior distribution is a joint Gaussian defined in terms of a mean $\mbf{m}_n$ over $\bm{\tau}$ and a corresponding covariance matrix $\bm{S}_n$ \citep{gp2006}.

This approach is most straightforward to apply in the univariate case where $\mbf{s}_n=(\mbf{t}_n,\mbf{x}_n)$. The model is defined in terms of a covariance function $\kappa_{\theta}(\cdot, \cdot)$ with parameters $\theta$. Unlike the case of basic kernel smoothing interpolation methods mentioned above, GPR models require the covariance function to be a valid Mercer kernel \citep{gp2006}. The GPR model also includes a noise variance term $\sigma^2$ and a mean function $\mu_{\phi}()$. Under this model, the posterior probability density over the values  $\mbf{x}'_n=[x'_{1n},...,x'_{Kn}]$ is given by $p(\mbf{x}'_n|\tau,\mbf{s}_n,\theta,\phi,\sigma)=\mathcal{N}(\mbf{x}'_n;\mbf{m}_n,\mbf{S}_n)$ where the conditional mean and covariance matrices $\mbf{m}_n$ and $\mbf{S}_n$ are shown below. $\mbf{K}_{\mbf{t}, \mbf{t}'}$ denotes the covariance matrix defined by $[\mbf{K}_{\mbf{t}, \mbf{t}'}]_{ij} = \kappa_{\theta}(\mbf{t}_i, \mbf{t}'_j)$. 
\begin{align}
    \bm{m}_n &= \mu_{\phi}(\bm{\tau})+\mbf{K}_{\bm{\tau}, \mbf{t}_n}(\mbf{K}_{\mbf{t}_n, \mbf{t}_n} + \sigma^2 \mbf{I})^{-1} 
    (\mbf{x}_n-\mu_{\phi}(\mbf{t}_n) )\\
    \bm{S}_n &=  \mbf{K}_{\bm{\tau}, \bm{\tau}} - \mbf{K}_{\bm{\tau}, \mbf{t}_n}(\mbf{K}_{\mbf{t}_n, \mbf{t}_n} + \sigma^2 \mbf{I})^{-1} \mbf{K}_{\mbf{t}_n, \bm{\tau}}
\end{align}
One possible application of GPR to the interpolation problem is to define $\mbf{s}'_{n}=(\bm{\tau},\bm{m}_n)$. This approach simply uses the posterior mean on $\tau$ as the interpolated values. This approach has no particular advantage over kernel smoothing methods as it discards all posterior uncertainty. A potentially better approach is to draw $J$ samples of the form $\mbf{x}'_{nj} \sim \mathcal{N}(\mbf{x}'_n;\mbf{m}_n,\mbf{S}_n)$ for $1\leq j\leq J$ and to construct $J$ interpolated time series using $\mbf{s}'_{nj}=(\bm{\tau},\mbf{x}'_{nj})$. The ensemble of interpolants $\{\mbf{s}'_{nj}|1\leq j\leq J\}$ will  exhibit greater cross-sample variation in regions where there are few observations in $\mbf{s}_n$, reflecting natural uncertainty due to observation sparsity. Such an ensemble can be used in down-stream analyses in a way that is analogous to multiple imputations in a missing data problem \citep{little2014statistical}.

However, as we can see, applying GPR in this way can be quite expensive as the naive computation of the matrix inverse in the posterior mean $\mbf{m}_n$ and covariance matrix $\mbf{S}_n$ computations requires time cubic in $L_n$. While faster approximations are available for these computations as well as for drawing samples from the GPR posterior, GPR-based approaches are typically still much slower than  methods derived from kernel smoothing. We also note that as with kernel smoothing methods, GPR has a number of parameters to estimate. These can be fit in a number of ways including the use of marginal likelihood maximization when given a data set consisting of irregularly sampled time series \citep{gp2006}.

The discussion of GPR thus far has focused only on the case of univariate irregularly sampled time series. Like with deterministic interpolation methods, GPR-based methods can also be applied to the case of multivariate irregularly sampled time series. The two basic approaches are again to leverage the series-based view and produce a posterior distribution or set of samples over each dimension separately, or to define a joint covariance function that applies both over time and across dimensions. A full GPR model over all dimensions (leveraging a multi-task or multi-output Gaussian process) has the advantage of being able to leverage covariance structure across all dimensions while producing a posterior distribution for any individual dimension at any time point \citep{bonilla2008multi}. However, defining flexible covariance functions for multivariate problems can be quite challenging due to the requirement of positive definiteness. 

\subsubsection{Similarity}

The previous section discusses the use of similarity functions and kernels within individual irregularly sampled time series. These approaches essentially aim to interpolate a single time series by considering the local similarity of time points and/or dimensions. However, we can also consider applying similarity to pairs of multivariate irregularly sampled time series $\mbf{s}$ and $\mbf{s}'$, providing a similarity-based modeling primitive.

At a minimum, a similarity kernel between irregularly sampled time series $\mathcal{K}_{\theta}(\mbf{s}, \mbf{s}')$ needs to provide non-negative values and needs to encode the notion that when $\mbf{s}$ and  $\mbf{s}'$ are more ``similar" in some sense, $\mathcal{K}_{\theta}(\mbf{s}, \mbf{s}')$ takes larger values. Any such kernel function $\mathcal{K}_{\theta}(\mbf{s}, \mbf{s}')$ can then be used to solve machine learning tasks that can be formulated in terms of similarity between data points. Basic examples include K-nearest neighbor classification and regression methods \citep{knn}. If the function $\mathcal{K}_{\theta}(\mbf{s}, \mbf{s}')$ also satisfies the requirements of a Mercer kernel function, a wider set of classical kernel methods can be applied including support vector classification \citep{cortes1995}, support vector regression \citep{smola2004}, kernel ridge regression \citep{hastie01statisticallearning}, kernel logistic regression \citep{wahba1999}, kernel principle components analysis \citep{scholkopf1998}, etc. 

A number of kernels have been proposed for irregularly sampled time series, some of which are valid Mercer kernels. For example, \cite{Lu2008} proposed an approach for the univariate case that can be thought of as constructing a latent function conditioned on an observed time series $\mbf{s}$ using a temporal similarity kernel. The construction of this latent function is identical to the construction of the Gaussian process mean function described in the previous section. \cite{Lu2008} then show that a kernel between two such latent functions induced by two different irregularly sampled time series $\mbf{s}$ and $\mbf{s}'$ can be computed only in terms of the observed values and time points that define the two time series. This construction is quite elegant in that it avoids the need to materialize the latent functions at reference time points. However, since it measures the similarity between deterministic latent functions, it discards all uncertainty due to sparse sampling. Other approaches include kernels that consider alignment and warping when assessing similarity \citep{shimodaira2001, cuturi2011}.

\subsection{Recurrence}

\begin{figure}[t]
    \centering
    \includegraphics[width=0.4\textwidth]{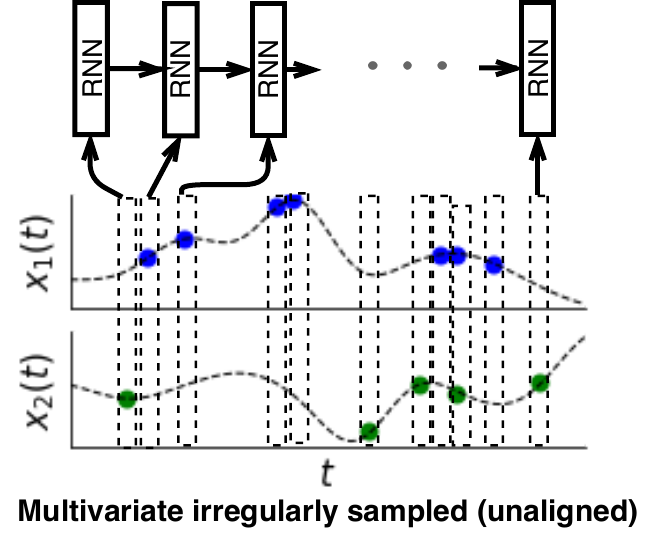}
    \caption{This figure illustrates the recurrence based modeling primitive for a  two-dimensional irregularly sampled time series with unaligned observations. In this case, we show the use of recurrent neural network (RNN) cell that integrates the input at each time point with the latent state from the previous time point. Basic RNN models view time series as general sequences and require completely observed input vectors. However, many RNN-based models have been developed that explicitly represent time to deal with irregular sampling and integrate methods for dealing with missing data.  }
    \label{fig:recurrence}
\end{figure}

As noted in the previous sections, one of the issues when dealing with time series in general is the possibility of different data cases containing different numbers of observations. This problem can be dealt with by defining models that appeal to primitives based on recurrence. 
The key property of such primitives is that they use a fixed, finite set of parameters to model sequences of arbitrary length. Examples include classical autoregressive structures used in Markov models, hidden Markov models \citep{hmm}, conditional random fields \citep{crf}, and recurrent neural networks \citep{rnn}. In this section, we will focus on recurrent neural network (RNN)  models as an example recurrent primitive  as shown in Figure \ref{fig:recurrence}. 

An RNN provides a basic building block for modeling fully observed multivariate sequences $\mbf{z}_n=[\mbf{z}_{1n},...,\mbf{z}_{L_nn}]$. The key is to define a cell whose hidden state $\mbf{h}_i$ is updated based on the previous state $\mbf{h}_{i-1}$ and the current input $\mbf{z}_{in}$ through a non-linear function $f_{\theta}()$ with parameters $\theta$. The model can also optionally output values $\hat{\mbf{y}}_{in}$ via a second function $q_{\phi}()$ applied to the hidden state. The model is thus able to process sequences of arbitrary length due to the fact that it processes elements of the sequence one at a time and the parameters $\theta$ and $\phi$ are position-independent.  
\begin{align}
    \mbf{h}_i &= f_{\theta}(\mbf{h}_{i-1}, \mbf{z}_{in})\\
    \hat{\mbf{y}}_{in} &= q_{\phi}(\mbf{h}_{i})
\end{align}
Due to the ability to process sequences of arbitrary length, RNNs are a useful building block for modeling time series in general. In the case of univariate irregularly sampled time series or multi-variate series with completely observed vector-valued observations, an RNN can be directly applied simply by processing the data in time order and discarding the values of the time points. Such an approach could be suitable for problems where the variation in inter-observation intervals is relatively small. When that is not the case, discarding the time values makes the model invariant to time gaps in a way that may be harmful to performance on tasks of interest. 

A number of simple approaches can be taken to solve this problem. One approach is to append the time points or inter-observation intervals to the vector-valued observations yielding the sequence of values $\mbf{x}'_{in} = [\mbf{x}_{in},t_{in}]$ or $\mbf{x}'_{in} = [\mbf{x}_{in},t_{in}-t_{i-1n}]$. An RNN model can then be applied to this modified sequence. In theory, a sufficiently powerful non-linear mapping $f_{\theta}()$ should be able to account for irregular sampling based on this representation. In practice, this approach can under-perform other approaches to dealing with time.

Recent work on ordinary differential equation (ODE) models \citep{neural_ode2018} in machine learning provides an alternative recurrence-based solution with better properties than traditional RNNs in terms of their ability to accomodate irregularly sampled data. In these models, ODEs are used to evolve the hidden state between continuous time observations. At each observation time-point, the hidden state is updated using a standard RNN update. These models are often referred to as ODE-RNNs. The hidden state update equations are defined below:
\begin{align}
\mbf{h}'_{i} &= \text{ODESolve}(g_\gamma, \mbf{h}_{i-1}, (t_{i-1n}, t_{in})) \\
\mbf{h}_i &= f_\theta(\mbf{h}'_{i}, \mbf{x}_{in})
\end{align}
The function $g_\gamma$ is a time-invariant function that takes the value  at the current time step and outputs a
gradient: $\frac{\partial \mbf{h}(t)}{\partial t} = g_{\gamma}(\mbf{h}(t))$. This function is parameterized using a neural network. The ODE defined by $g_{\gamma}(\mbf{h}(t))$ is then solved at the given time point $t_{in}$ to produce a new hidden state $\mbf{h'}_{i}$. This hidden state is further updated to fold in the input value at time $t_{in}$, yielding the final hidden state value $\mbf{h}_i$ at time $t_{in}$.
This model does not explicitly depend on $t_{in}$ or $t_{in}-t_{i-1n}$ when updating the hidden state, but does depend on time implicitly through the resulting latent state of the dynamical system. 

The ODE-RNN model primitive is a substantially more elegant approach to dealing with irregular sampling than adaptations of discrete time RNNs, but it can also be substantially slower due to the need to repeatedly apply an ODE solver. Further, we note that when irregularly sampled time series are multivariate, but the vector-valued observations are not fully observed, the application of both  standard RNN models and ODE-RNNs becomes significantly more challenging as both models require a fully observed $\mbf{x}_{in}$ vector in order to update the hidden state.

A common baseline approach is to perform some form of imputation by defining a new sequence of values $\mbf{x}'_{in} = \mbf{r}_{in}\; \mbf{x}_{in} + (1 - \mbf{r}_{in}) \;\Tilde{\mbf{x}}_{in}$ where $\Tilde{\mbf{x}}_{in}$ is the value to impute when the response indicator $\mbf{r}_{in}=0$. Forward filling, zero imputation, and mean imputation can sometimes be reasonable options when the volume of missing data is low.

\subsection{Attention}
\begin{figure}[t]
    \centering
    \includegraphics[width=0.4\textwidth]{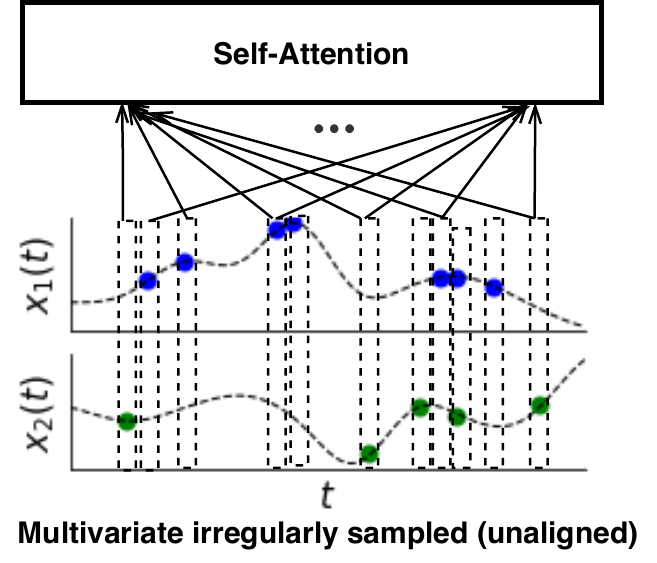}
    \caption{This figure illustrates the attention modeling primitive for two dimensional irregularly sampled time series.  The attention-based primitive processes time series in parallel instead of sequentially as for an RNN. Attention models learn which regions of an input time series to attend to when computing outputs at different points in time by leveraging positional  or time encodings.}
    \label{fig:attention}
\end{figure}

Attention models have become a key modeling component for many machine learning tasks including image caption generation, speech recognition and neural translation \citep{transformer}.  In self-attention models, each element of an output sequence learns which  elements of an input sequence to attend to without relying on recurrent network structures. Self-attention models offer computational advantages over RNNs since sequence processing can be fully parallelized. Self-attention is a particular instance of scaled dot-product attention which is defined as:

\begin{align}
    \text{Attn}(\mbf{Q}, \mbf{K}, \mbf{V}) = \text{softmax} \bigg( \frac{\mbf{Q}\, \mbf{K}^T}{\sqrt{C}}\bigg) \mbf{V}
\end{align}
where $\mbf{Q}, \mbf{K}, \mbf{V}$ denote the query, key, value representation  respectively, and $C$ is the dimension of the key or query representation. In self-attention, all of the queries, keys and values come from the input sequence. Self-attention relies on positional encoding, a vector representation for each position in the sequence to recognize and capture the sequential ordering information. The above defined $\mbf{Q}, \mbf{K}, \mbf{V}$ matrices are typically linear projections of position-value joint representations. 

To provide a primitive for irregular sampling (as shown in Figure \ref{fig:attention}), time values can be converted into a vector representation similar to positional encoding and concatenated with the observation value as described for RNNs. Missing values in vector-valued observations are also problematic for attention-based modules, which (like standard RNNs) expect fully observed vectors as input. Approaches based on attention thus also need to solve the problem of missing dimensions, and a variety of imputation solutions can be used as described previously. 

\subsection{Structural Invariance}
\label{sec:set}

While the modeling primitives presented to this point all attempt to represent the time ordering of observations in the structure of the model this is not strictly required. Leveraging such structural invariance has the potential to greatly simplify learning problems. 

\begin{figure}[t]
    \centering
    \includegraphics[width=\textwidth]{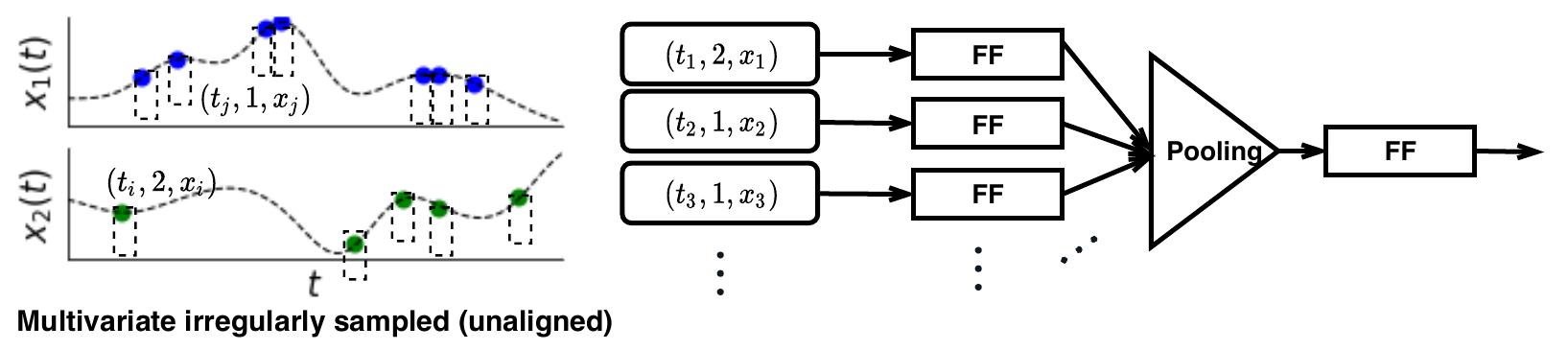}
    \caption{An illustration of the structural invariance-based modeling primitive. Inspired by the set-based view of a multivariate irregularly sampled time series, this approach processes individual (time, value, dimension) tuples via an encoding function and then pools over the output of all such tuples. The model structure thus does not reflect the time ordering of the data in any way.}
    \label{fig:invariance}
\end{figure}

Recently, set-based neural network approaches (as shown in Figure \ref{fig:invariance}) have been employed to deal with irregularly sampled data and they exactly express this form of structural invariance. Set-based model architectures \citep{deepsets} support variable length sequences, partially observed vectors, sparse observations and irregular intervals between observation times while avoiding the need for temporal discretization, imputation and interpolation. These approaches leverage the set-based representation of an irregularly sampled time series data set $\mbf{s}_n = \{(t_{in},d_{in},x_{in})| 1\leq i\leq L_n\}$. As shown below, such approaches produce an encoding of an input irregularly sampled time series by applying an initial encoder $f_\theta$ to individual time-dimension-value tuples. 
\begin{align}
    \mbf{h} = g_\phi(\mbox{pool}(\{f_\theta(t_{in},d_{in},x_{in})| 1\leq i\leq L_n\})
\end{align}
The approach then performs a pooling operation over all initial encodings in a way that is completely invariant to the temporal structure of the data. The output of pooling is mapped through one additional set of encoding layers  $g_\phi$ to produce the final representation. Commonly used pooling functions are max, mean and sum.

\subsection{Summary} The discretization, interpolation, similarity, recurrence, attention and structural invariance modeling primitives that we have just described underlie essentially all of the recent work on models for irregularly sampled time series across all of the tasks described in Section \ref{tasks}. In the following sections, we survey over 40 papers categorized by the fundamental modeling primitive used and discuss the tasks to which each method can be applied.

%% file: discretization.tex
\section{Discretization-Based Approaches}
\label{sec:discretization_approaches}

In this section, we discuss approaches to modeling irregularly sampled time series that leverage discretization as their primary modeling primitive for accommodating irregular sampling.
\citet{marlin-ihi2012}, \citet{lipton2016directly}, \citet{bahadori2019} and \citet{benchmark} all apply discretization as a modeling primitive for dealing with irregular sampling. These papers all discretize time into consecutive, hour-long, non-overlapping intervals within a fixed time interval $[0,T]$ common to all data cases. 
The application of discretization under these conditions enables the use of models that operate on fixed-dimensional vectors. However, in all of the tasks considered in these papers, the application of discretization results in missing data that must be dealt with, as well as instances where aggregation is required within time intervals.

\citet{marlin-ihi2012} make the assumption that the missing data is missing at random and apply probabilistic mixture models that can efficiently deal with missing data under this assumption. This allows models to be learned without requiring explicit imputation. \citet{marlin-ihi2012} leverage this capability of the probabilistic mixture model framework to define a generative mixture of experts classifier for whole time series classification. The case of multiple values for a given dimension being defined within the same time interval is dealt with via averaging.

\citet{lipton2016directly} apply discretization followed by an RNN to build a whole time series classification model. This approach requires explicit imputation of missing data created during the discretization step. They consider two basic approaches: forward-filling and zero imputation. Missing values are replaced with zeros in the zero-imputation strategy. In the forward-filling approach, a missing value $\mbf{x}_{idn}$ at time $t_{in}$ on dimension $d$ is set to the last observed value on dimension $d$ (e.g., the value of  $\mbf{x}_{jdn}$ where $j$ is
such that $t_{jn}$ is the largest time value satisfying  $t_{jn}<t_{in}$ and $\mbf{r}_{jdn}=1$). In the absence of previous measurements (or if the variable is missing entirely), it is replaced with the-median estimated over all measurements in the training data.   

One important feature of missing data problems is the potential for the sequence of observation times to itself be informative \citep{little2014statistical}. Since the set of response  indicators $\mbf{r}_{idn}$ is always observed, this information is easy to condition on as well. 
This approach was also used by \citet{lipton2016directly} where they include as input to the RNN both  the observed and imputed values of the multivariate time series and the values of the response indicator vector. In addition to the binary indicator variable, \citet{lipton2016directly} add hand-engineered features derived from the response indicator time series such as mean and standard deviation of indicator variables for each time series.

\citet{benchmark} also consider the application of discretization followed by RNNs. They apply averaging or selection of the last time point in an interval to deal with multiple observations in the discretization window, and use  mean imputation or forward filling to deal with missing values. \citet{benchmark} augment  standard RNN models by predicting outputs for multiple tasks jointly. Unlike in  standard RNNs where the hidden state from the last position in the sequence is decoded to predict supervised outputs, \citet{benchmark} apply supervision at each time step. This framework has been applied to whole time series classification
tasks, detection tasks and prediction tasks.

\citet{attendanddiagnose2018} follow \citet{benchmark}\footnote{\citet{benchmark} had an archive version available since 2017.} and also discretize the sparse and irregularly sampled time-series data into hour long intervals, and use similar imputations for dealing with missing data. 
This model has been applied to the tasks of whole time series classification, detection and prediction.
However, \citet{attendanddiagnose2018}  adopt the multi-head attention mechanism similar to \citet{transformer} instead of an RNN as the primary model structure. They use positional encoding to incorporate the temporal order into the representation learning. They use the dense interpolation technique \citep{trask2015} to obtain a unified representation for a sequence.

Several approaches have also employed an encoder-decoder framework for learning with missing data in time series, which can also be applied to irregularly sampled time series after discretization.  
\citet{Bianchi2018} proposed an autoencoder-based approach to learn representations of multivariate time series with missing data for the whole time series classification and smoothing tasks. They use a standard RNN as the decoder. For the encoder, they use a bi-directional RNN where they combine the hidden layer output of forward and backward RNNs using a fully-connected layer. In the presence of missing data, they replace the missing values with zero or the mean while encoding, and use the decoder as a final imputer. \citet{Bianchi2018} also introduce a kernel alignment procedure to preserve the pairwise similarities of the inputs in the learned representations. These pairwise similarities are encoded in a positive semi-definite matrix that is also passed as input to the model. \citet{fortuin20} proposed a variational autoencoder (VAE)  \citep{kingma13, rezende14} approach for the task of smoothing in multivariate time series with a Gaussian process prior in the latent space to capture temporal dynamics.


In summary, temporal discretization provides a solution for learning from irregularly sampled time series that replaces the problem of irregular observation intervals with the problem of missing data. While convenient in some respects, this approach requires several ad-hoc choices including the width of the discretization windows and the aggregation function used within windows. In the next section, we turn to methods based on the use of interpolation as a modeling primitive. These approaches still require specifying a discrete set of interpolation points, but use more sophisticated global methods for defining values at the interpolation points.

%% file: interpolation.tex
\section{Interpolation-Based Approaches}
In this section we discuss methods for learning from irregularly sampled time series that are based on interpolation as a modeling primitive. Similar to discretization methods, interpolation methods require specifying a discrete set of reference time points. However,  they materialize interpolants at these time points in a way that can leverage all available observations in the input if desired. This can make them much more powerful than discretization methods with ad-hoc local aggregation functions. 


To begin, \citet{Shukla2019} proposed the interpolation-prediction network (IPNet) framework, which uses  several semi-parametric radial basis function (RBF) interpolation layers to produce multiple interpolants given an irregularly sampled multivariate  time series as input. This framework leverages the series-based view of an irregularly sampled time series. The first layer of the interpolation network separately transforms each of the $D$ dimensions, creating several intermediate interpolants. The second interpolation layer merges information across all dimensions at each reference time point by taking into account learnable correlations across all dimensions. The prediction network takes the output of the interpolation network as its input and produces a prediction. The prediction network can consist of any standard supervised neural network architecture (fully-connected feed forward, convolutional, recurrent, etc.). This model has been applied to the tasks of whole time series regression and classification.

One of the drawbacks of the IPNet framework is that it uses fixed temporal RBF kernels with adjustable parameters. This biases the model to only being able to express certain notions of temporal similarity. \citet{shukla2020multitime} present a related continuous-time interpolation-based approach that replaces the fixed RBF kernels in the IPNet framework with learnable temporal similarity functions constructed using learnable embeddings of continuous time values and a temporal attention mechanism \citep{transformer}. This added flexibility appears to lead to improved predictive performance. This framework has been applied to the interpolation task, whole time series classification task, and detection task.  

\citet{Li2020} proposed an encoder-decoder architecture for modeling irregularly sampled time
series data. The encoder in this framework is based on a piecewise linear function, which learns an interpolation on a fixed set of reference time points. This framework uses a kernel smoother as a decoder to interpolate at arbitrary
times. 

%

One of the potential disadvantages of using deterministic kernel smoothing-based interpolation approaches such as IPNets is that they do not reflect uncertainty due to long intervals with no observations. An alternative line of work on interpolation-based approaches instead leverages Gaussian processes as a building block, enabling uncertainty propagation as described previously \citep{gp2006}. In response to the shortcomings of the kernel-based approach of \citet{Lu2008}, \citet{li2015classification} developed a two-step interpolation-based approach to modeling the similarity between sparse and irregularly sampled time series in a way that is sensitive to uncertainty. In the first step of this approach, they use marginal likelihood maximization to fit a Gaussian process regression model to a data set of irregularly sampled time series. Once the model is trained, they define a set of evenly spaced interpolation points and materialize the Gaussian process posterior for each time series at the same set of reference time points. In the second phase of the method, they define Mercer kernels between the  Gaussian process posteriors at the reference time points. This approach allows for the use of irregularly sampled time series with classical supervised and unsupervised kernel methods such as support vector machines and kernel PCA. 



In subsequent work, \citet{li2016scalable} showed how the materialized Gaussian process regression posterior representation could be extended for use as an uncertainty preserving interface between irregularly sampled time series and arbitrary deep learning model components that expect regularly spaced or fixed dimensional inputs. Further, this work showed how the Gaussian process regression model parameters and the deep learning model parameters can be learned jointly and end-to-end using the re-parameterization trick \citep{kingma2015variational}. This model structure allows for solving a variety of tasks, but was primarily evaluated in the context of whole time series classification.

While the model of \citet{li2016scalable}
was only applied to univariate irregularly sampled time series in the original work, in follow-up work the model was extended to multivariate time series using a multi-output
Gaussian process regression model \citep{futoma2017improved}. 
However, modeling multivariate time series within this framework is quite challenging due to the constraints on the covariance function used in the GP regression layer. \citet{futoma2017improved} deal with this problem using a sum of separable kernel functions \citep{bonilla2008multi}, which somewhat limits the expressiveness of the model. The work of \citet{futoma2017improved} was also focused on the whole time series classification as a task. 

In terms of the tasks of interpolation and forecasting, \citet{Ghassemi2015} proposed a multi-task Gaussian process (MTGP) model for irregularly sampled physiological signals and clinical notes, which jointly transforms the measurements into a unified latent space. They showed that MTGP models provide better performance than single task GP models for interpolation and forecasting.  \citet{Liu2016LearningAF} combine state-space models with GPs for learning  forecasting models from irregularly sampled multivariate clinical data. \citet{soleimani2017} also employed multi-output GPs to jointly model multivariate physiological signals. 


In summary, interpolation as a modeling primitive provides more sophisticated approaches than the more basic discretization-based methods. The two main families of approaches explored to date in this area are deterministic kernel smoothing methods such as IPNets and probabilistic Gaussian process-based methods. Both approaches naturally accommodate continuous time observations and can provide interfaces to other modeling building blocks such as kernel machines and neural networks. The main advantage of Gaussian process-based modeling primitives is the ability to represent and propagate uncertainty into downstream model components. Their main drawback is the significantly higher run times during both training and prediction, and the added complexity of needing to define valid mercer kernels for multivariate time series. Deterministic interpolation frameworks like IPNets do not have the elegance of Gaussian processes in terms of representing uncertainty, but can be much more flexible in terms of how they express temporal and cross-dimensional relationships. Importantly, they are also one to two orders of magnitude more computationally efficient during training.  In the next section we turn to the case of recurrence as a modeling primitive and explore methods based on recurrent neural networks.

%% file: recurrence.tex
\section{Recurrence-Based Approaches}

In this section, we discuss approaches to modeling irregularly sampled time series that leverage recurrence as their primary modeling primitive for accommodating irregular sampling. We begin with traditional discrete RNN models, and then discuss more recent work on ODE-based models. Importantly, methods in this section do not rely on discretization or interpolation prior to the application of RNNs. This is the primary distinction relative to methods presented in the previous sections, some of which also leverage RNNs after discretization and interpolation have been applied to accommodate irregular sampling. 

\subsection{RNN-Based Methods}


Similar to \citet{lipton2016directly}, \citet{che2016recurrent}
present several methods based on  gated recurrent unit networks
(GRUs, \citet{gru}) for the whole time series classification task. Their approach is cast in the vector-based representation. Thus, even though they do not apply discretization, they still need to deal with the problem of missing data inherited from this view of the problem. In their baseline methods, \citet{che2016recurrent} deal with the issue of missing values in vectors observed at irregularly sampled time points using simple imputation methods including mean imputation and forward filling. They also consider augmenting inputs with binary response indicator values and per-dimension inter-observation time intervals. This last approach attempts to provide the RNN with information about how long it has been since a value on a given dimension was last observed. When applying forward filling as the imputation method, the inter-observation time interval values indicate how long it has been since a new value for the variable was last observed. 

However, the primary contribution of \citet{che2016recurrent} is the  GRU with decay (GRU-D) model that adds a temporal decay mechanism to the the input variables and the hidden states.
The decay rate function is defined as:
\begin{align}
    \gamma(\delta) = \exp \{-\max (0, \mbf{W}_\gamma \delta + \mbf{b}_\gamma) \}
    \label{eq:decay}
\end{align}
where $\mbf{W}_\gamma$ and  $\mbf{b}_\gamma$ are learned jointly with GRU parameters at training time. $\delta$ represents the length of the inter-observation interval. They introduce input decay for missing variables to decay the previously observed value toward the empirical mean for a given dimension over time. For large inter-observation intervals, this can be a more sensible choice than continued forward filling as the previously observed value will become uncorrelated with the current value over sufficiently long intervals. This can make the overall mean for that dimension a better imputed value. We describe the GRU-D layer below: 
\begin{align}
    \mbf{x}'_{idn} = \mbf{r}_{idn}\; \mbf{x}_{idn} + (1 - \mbf{r}_{idn}) \;(\gamma(\delta) \mbf{x}_{jdn} + (1 - \gamma(\delta)) \bar{\mbf{x}}_d) 
    \label{eq:imputed}
\end{align}
where $\mbf{x}_{jdn}$ is the last observation $(t_j < t_i)$ for dimension $d$, and $\bar{\mbf{x}}_d$ is the empirical mean of the $d^{th}$ dimension in multivariate time series.  Importantly, while this approach can be thought of as a form of imputation, the parameters that control imputation are learned end-to-end with the rest of the model parameters based on a supervised objective. 

As a further modification, \citet{che2016recurrent} also consider decaying the hidden states in the absence of observations. This has the effect of partially resetting the hidden state when there is a long interval with no observations. This is also a sensible modification as any accumulated hidden state may be irrelevant following a long time interval with no observations. \citet{Choi2015} employed a similar method where they concatenate the time gaps between observations with the actual observations and supply it directly as input to a GRU model. On the other hand, \citet{pham2016} proposed to capture time irregularity by modifying the forget gate a LSTM \citep{lstm} using a decay combined with a parametric function of the time gap between current and past measurement. 

\citet{Neil} proposed the phased LSTM for supervised learning from irregularly sampled time series. The phased LSTM model introduced a time-dependent gating mechanism by adding a new time gate which regulates access to the hidden and cell state of the LSTM. The time-aware LSTM (T-LSTM) \citep{baytas17} instead transforms time intervals into weights using a decay function and uses them to adjust the hidden state passed from the previous time step.
While these approaches allow the network to handle event-based sequences with irregularly spaced observations, they do not support incomplete multivariate observations.

\citet{Li2019} argue that the time difference $\delta$ between the current time step and last observation used in the dynamic imputation mechanism is not sufficient for the model to fully capture the  overall pattern of missingness across a time series. They present VS-GRU, which also considers the observation rate of each dimension.
They adopt a similar decay mechanism to that used by GRU-D. \citet{Li2019} also proposed another variant VS-GRU-i that adds a standard GRU layer on top of VS-GRU. They add a penalized mechanism at the input of the second GRU layer to detect if one variable is completely missing or there are only a few observed values. 

\citet{Kim2018} also propose an approach to dealing with missing data in an RNN using decay, which they refer to as a temporal belief memory. When an input is missing, the belief gate computes the belief of the last observation based on the time gap between the current time and the last observation time. If the belief is greater than a threshold, the model imputes using the last observation; otherwise, it sets the missing value to the mean value of observations for that dimension. The belief gate function $\gamma(\delta)$ for a given dimension $d$ is defined as:

\begin{align}
    \gamma(\delta) &= \begin{cases}
    1 & \text{if } \exp(-\beta_d \delta/ \tau) > \pi\\
    0 & \text{otherwise} 
    \end{cases}
\end{align}
%
where $\beta_d$ is a tuneable decay parameter, $\tau$ is a window length, and $\delta$ is the time difference from the last observation to  the current observation. Imputed input can then be defined using Equation \ref{eq:imputed}.





One drawback of all of the approaches mentioned so far is that they process an input time series in time order. This is clearly a disadvantage for RNN-based models that attempt to accommodate for missing values under a vector-based representation as any imputation that occurs corresponds to the application of a filtering approach. This is overly restrictive when models are applied to problems like interpolation or whole time-series classification as in these problems it is permissible to integrate information from both the relative past and relative future of each time point. Indeed, methods like GRU-D make maximally uninformative imputations immediately prior to the next observation, whereas smoothing approaches would be informed by the next as well as the previously observed values in the time series.   

A separate line of work has looked at using data from future as well as from the past for imputation of  missing data in RNNs. \citet{Yoon2017} and  \citet{Yoon18} present an approach based on a multi-directional RNN (M-RNN) which operates across dimensions as well as within dimensions. It consists of two separate blocks for imputing  missing values. They first construct an imputation function using a bi-directional RNN that operates within a given dimension. The second imputation block constructs an imputation function using a fully connected layer that operates only across streams at a given time point. They also concatenate the input with the response indicator vector and time gap from the last observed value to the current timestamp. 

\citet{brits2018} also proposed models for imputing the missing values for multiple correlated time series using smoothing-like methods using RNNs. Their basic approach learns the missing values directly in a recurrent dynamical system based on the observed data. They combine the hidden decay mechanism of GRU-D \citep{che2016recurrent} and the RNN missing data imputation methods proposed in \citet{Kim2017}. 
They also propose a bidirectional variant, BRITS-I, where they model the recurrent dynamics in the forward as well as backward direction. They introduce an additional loss in the bidirectional case to enforce consistency of predictions obtained in both directions. They extend BRITS-I for correlated recurrent imputation where the imputation is a weighted sum of history-based estimation and feature-based estimation. History-based estimation is simply uncorrelated recurrent imputation while feature-based estimation is computed based on only other features available at that time. They also concatenate the input with mask variables and time gaps from the last observed value to the current time point.


\citet{datagru2020} also introduce a time-aware structure based on a GRU to handle irregular time intervals. Similar to GRU-D, they handle irregular sampling by decaying the hidden state between the successive observations. However, instead of using exponential decay as in GRU-D, they apply inverse log decay, which they claim works better:
\begin{align}
    \hat{\mbf{h}}_{t-1} = {\frac{1}{\log (e + \delta_t)}} \odot \mbf{h}_{t-1}
\end{align}
\citet{datagru2020} use a dual attention mechanism for dealing with missing values caused by partially observed vectors or misalignment of observation time points on different dimensions. They perform imputations using Gaussian processes and a separate time-aware GRU model and combine them using an embedding layer to get the transformed input.

\citet{Luo2018} and \citet{Guo2019} employed generative adversarial networks (GANs) to impute missing values in irregularly sampled time series data by leveraging  GRU-D style models to take into account irregular sampling. They propose a 2-stage method for imputation of missing data in time series. In the first stage, they train a GAN with GRU-D as generator and discriminator.  In the second stage, they train the input ``noise” of the generator of the GAN so that the generated time series is as close as possible to the original incomplete time series and the generated data has high probability of being real. They achieve this by minimizing a weighted sum of reconstruction loss and discriminative loss. 

\citet{Luo2019} propose a single-stage generative model
to impute missing values in multivariate time series without the need of the two-stage process as in \citet{Luo2018}. They use a two-layer GRU-D network \citep{che2016recurrent} and a fully connected layer as generator. They begin with a zero imputed input. After a recurrent processing of the input time series using GRU-D, the last hidden state is processed using a fully connected layer which outputs a compressed low-dimensional vector. The compressed low dimensional vector acts  as the input to another fully connected layer which transforms it into the initial input for the second GRU-D layer. The second GRU-D layer then outputs the generated sample at every time step. The generator model is trained in a denoising autoencoder fashion where they drop out original samples and reconstruct them. Since it is not a good idea to drop out samples when the missing rates are very high, they add some noise to the input samples and train a  denoising autoencoder. The discriminator model is implemented using a GRU-D layer  and a fully-connected layer. The fully connected layer takes the hidden state of GRU-D and outputs the probability of being true. The task of the discriminator is to distinguish between generated sample and the input/true sample. 

\citet{Che2018} proposed a deep generative model that uses a latent hierarchical structure to capture the temporal dependencies of multivariate time series with different sampling rates. This model captures the underlying data generation process by using a VAE-based approach and learns latent hierarchical structures using learnable switches and auxiliary connections. In particular, the switches use an update-and-reuse mechanism to control the updates of the latent states of a layer based on their previous states and the lower latent layers. The auxiliary connections between time series of different sampling rates and different latent layers help the model effectively capture short-term and long-term temporal dependencies. Their inference network consists of different RNN models for time series with different sampling rates. They jointly learn the parameters of the generative model and the inference network by maximizing the ELBO. In the case of irregularly sampled time series or missing data, missing values are interpolated by adding auxiliary connections using a model trained using only the observed data points. While training, missing data points are replaced with zero in the inference network and the corresponding auxiliary connections are removed in the generative model.

In summary, a great deal of work in the area of modeling and learning from irregularly sampled time series has been based on modifications of standard recurrent neural network models. RNN-based approaches have the primary advantage that they can deal with input sequences of different lengths. However, basic RNNs have no direct ability to deal with irregular sampling and no way to deal with missing values. GRU-D and related methods attempt to solve both problems using temporal decay, but the resulting models lack true continuous time semantics. In the next section, we turn to ODE-based recurrent models, which can provide a more elegant approach to modeling continuous time data at increased computational cost.

\subsection{ODE Based Methods}

\citet{neural_ode2018} proposed a variational auto-encoder (VAE) model \citep{kingma13, rezende14} for continuous time data based on the use of a neural network decoder combined with a latent ordinary differential equation (ODE) model. They model time series data via a latent continuous-time function $\mbf{z}(t)$. 
This function is defined via a neural network representation of its gradient field that takes the form $\frac{\partial \mbf{z}(t)}{\partial t} = f_{\theta}(\mbf{z}(t))$. The full generative process for a single data case sampled from the neural ODE latent time series model is shown below. Note that the generative process conditions on the set of observation time points $\mbf{t}=t_0,...,t_L$. 
\begin{align}
\mbf{z}(t_0) &\sim p(\mbf{z}(t_0)) \\
\mbf{z}(t_1), \cdots, \mbf{z}(t_L) &= \text{ODESolve}(\mbf{z}(t_0), f, \theta, t_0, \cdots, t_L) \\
\label{eq:ode_solve}
\mbf{x}_i &\sim p(\mbf{x}_i | g_{\phi}(\mbf{z}(t_i)))
\end{align}
While this model has many appealing properties as a continuous time generative model, VAE's also require the specification of a recognition network or encoder function to map observations into the latent space. Due to the deterministic nature of the ODE model, it suffices to identify the distribution over the latent space at any single time $t$. \citet{neural_ode2018} use a basic RNN model as the encoder/recognition network for computing the approximate posterior. In the case of irregularly sampled time series, \citet{neural_ode2018} propose to apply the encoder to the sequence of observed values backward in time from the end of the time series to the beginning.

However, the RNN encoder used did not account for variable inter-observation intervals at all and also did not account for partially observed input vectors. This means that the overall model has fairly asymmetric capabilities with the ODE used only in the generator and much more basic components applied in the encoder. \citet{Rubanova2019} subsequently proposed using an ODE-RNN model as an encoder for the latent ODE model. This yields a much more capable model with the intrinsic ability to accommodate irregularly sampled time series of fully observed vectors. 

However, this model still has the limitation that it does not directly accommodate irregularly sampled time series of incompletely observed vectors. In addition, the encoder used by \citet{Rubanova2019} is also applied backwards in time only. While in theory all information in the input time series can be encoded into the latent state at the start of the time series, in practice this structure can be limiting as the encoder may not be able to adequately preserve information about structures in later sections of the time series. 

\citet{DeBrouwer2019} proposed a related method based on a combination of ODE and GRU components that they refer to as GRU-ODE-Bayes. This model is a  continuous-time version of the Gated Recurrent Unit model \citep{gru} that can be applied to irregularly sampled time series. Instead of the encoder-decoder architecture where the ODE is decoupled from the input processing, GRU-ODE-Bayes provides a tighter integration by interleaving the ODE and the input processing steps. GRU-ODE-Bayes uses the ODE to propagate the hidden state between continuous-time observations. It updates the current hidden state whenever a new observation is available. 

Contrary to the previous approaches in this subsection, GRU-ODE-Bayes can also handle partially observed vectors, which are common in irregularly sampled time series. Partially observed input in the RNN hidden update equation is replaced with a transformed input defined as a function (with learnable parameters) of previous hidden state, partially observed input and mask variable, similar to the approach used in the GRU-D model. However, due to the structure of the GRU-ODE-Bayes model, it only has access to information from the relative past of individual time points when accounting for missing data. This means that it suffers from the same problem as the GRU-D model where imputed values for missing dimensions have maximum uncertainty immediately before the next observation. This makes GRU-ODE-Bayes a poor choice for a smoothing or interpolation method, while the local structure of the hidden state updates make it better suited for a filtering tasks. The final hidden states can also be used as input to prediction and detection tasks.

One of the disadvantages of the latent ODE models defined by Equation \ref{eq:ode_solve} is that once $\theta$ has been learned, then the solution of Equation \ref{eq:ode_solve}  is determined by the initial condition at $z_0$, and there is no direct mechanism for incorporating data or hidden state at a later time. ODE-RNN tries to solve this problem by using a RNN in conjunction with ODE which updates the hidden state in the presence of a new observation and then solves the ODE using this new hidden state as the initial point. This leads to jumps in the hidden state, which is not suitable for defining continuous time dynamics.  \citet{kidger2020} solve this problem by defining Neural Controlled Differential Equation (CDE). Neural CDEs are capable of processing incoming irregularly sampled data in a more principled way. Neural CDEs also have the capability to deal with incomplete observations by independently interpolating each channels. 

In summary, ODE-based modeling primitives provide an interesting basis for learning from continuous time data. They solve the problem of accommodating irregularly sampled time series of fully observed vectors in a way that is significantly cleaner than standard RNNs. However, ODE models can be quite a bit slower during training and deployment due to the need to make repeated calls to an ODE solver. Finally, some of the model structures proposed to date are also limiting in their inability to cleanly accommodate incomplete observations. While Neural CDEs are a very recent approach, they appear to have interesting advantages over the ODE-RNN framework and related models by providing an improved approach to integrating observations through time.

%% file: attention.tex
\section{Attention-Based Approaches}
Several recent models have leveraged attention mechanisms as their fundamental approach to dealing with irregular sampling \citep{seft,attendanddiagnose2018,attain2019}.
Most of these approaches are similar to  \citet{transformer} where they replace the positional encoding with an encoding of time and model sequences using self-attention. Instead of adding the time encoding to the input representation as in \citet{transformer}, they concatenate it with the input representation. These methods use a fixed time encoding similar to the positional encoding of \citet{transformer}. \citet{selfatttimeemb2019}  learn a functional time representation and concatenate it with the input event embedding to model time-event interactions. 

\citet{attain2019} proposed an attention-based
time-aware approach that models the time irregularity between events by employing the attention mechanism. This approach adjusts the memory of an LSTM when accumulating previous information.  Instead of reading the information from just one previous cell state, it combines values in previous cell states using attention weights and time gaps.

\citet{Choi2016} learn an interpretable representation of irregularly sampled events using a two-level attention model. Attention vectors are generated by running RNNs backwards in time. These models are subsequently used to compute a final representation using a weighted average of the event embedding with the corresponding attention weights. They showed that performance can be improved by concatenating the corresponding timestamps with the event embedding.

\citet{selfatttimeemb2019} proposed a set of time embedding methods for functional time representation learning, and demonstrate their effectiveness when combined with self-attention in continuous-time event sequence prediction. The proposed functional forms are motivated from Bochner's \citep{bochner} and Mercer's \citep{mercer} theorem.

\citet{seft} employed a transformer-based \citep{transformer}  approach for modeling irregularly sampled time series. This approach deals with irregular sampling by defining a time embedding to represent the time point of an observation.  The time embedding defined here is a variant of positional encoding \citep{transformer} which takes continuous time values as input. The transformer architecture
was applied to the time series by concatenating the vectors of each time point with a mask variable indicating whether the variable is observed or missing and the corresponding time embedding. If the observation is missing, the input is set to zero for that dimension. The output of the transformer architecture is a sequence of embeddings, which can be aggregated into a final representation using mean-pooling for classification tasks. 

%% file: set.tex
\section{Structural Invariance-Based Approaches}
\label{sec:invariance_approaches}
In this section we consider work in the area of using structural invariance as a modeling primitive. Such approaches leverage the set-based view of an irregularly sampled time series to build a model whose structure does not depend on the time order of the input time series at all. 

In recent work, \citet{seft} propose a set function-based approach for the task of classification and detection on time-series with irregularly sampled and unaligned observations. This approach can inherently handle irregular sampling and partially observed vectors. Similar to the positional encoding used in transformer models \citep{transformer}, \citet{seft} define an embedding for continuous values of time. This approach represents the irregularly sampled time series data as a set of tuples consisting of a time embedding, value and dimension indicator to differentiate between different dimensions. Following the approach described in Section \ref{sec:set}, tuples are independently transformed using a multi-layer feed-forward network. \citet{seft} suggest a
weighted mean approach to aggregate the encoded tuples in order to allow the model to decide which
observations are relevant and which should be considered
irrelevant. This is achieved by computing attention weights
over the set of input elements, and subsequently, computing
the weighted sum over all elements in the set corresponding to their attention weights.

%% file: performance.tex
\section{Predictive Performance}
\label{sec:performance}
In this section, we summarize evidence regarding their relative predictive performance of methods presented in the previous sections. We provide a brief introduction to the data sets that are commonly used to evaluate models for sparse and irregularly sampled time series in Appendix \ref{sec:datasets}. In Appendix \ref{sec:code}, we provide links to open source implementations of many of the methods described in this and earlier sections.

In terms of discretization-based methods, \citet{lipton2016directly} showed that discretization combined with zero imputation and conditioning on response indicators outperformed LSTM models that do not condition on response indicators. \citet{lipton2016directly} also showed that in presence of missing indicators, zero imputation performed better than forward filling based imputation. \citet{benchmark}  demonstrated the advantages of using channel-wise LSTMs and learning to predict multiple tasks using a single neural model. \citet{attendanddiagnose2018} showed that the performance of \citet{benchmark}  on the challenging
MIMIC-III benchmark data set can be further improved by using self-attention instead of an LSTM. \citet{bahadori2019} showed that a temporal clustering based approach achieves better performance than \citet{attendanddiagnose2018} and \citet{benchmark} on classification and detection tasks.

However, discretization-based methods for irregular sampling data are typically outperformed by models with the ability to directly use an irregularly sampled time series as input. 
\citet{Kim2018} showed that a simple imputation based on the time gap between the current time and the last observed time achieved better performance than imputation methods based on mean and forward-filling on classification tasks. They also showed that using response indicators further improved  performance.
\citet{Choi2016} showed that the classification performance of their attention model on irregularly sampled events can be improved by concatenating the corresponding timestamps
with the event embedding. 

\citet{che2016recurrent} and \citet{pham2016} showed that introducing a decay function based on the time gap between current and past measurement inside a RNN model improves performance on classification and prediction tasks as compared to several off-the-shelf machine learning models and RNN baselines based on mean and forward filling imputation schemes as well as RNN models based on concatenating time gaps and missing indicators. \citet{che2016recurrent} also showed that for classification and prediction tasks, combining exponential decay of hidden states with an imputation scheme based on the weighted average of previous values and the empirical mean  further outperforms RNNs that use exponential decay on hidden state only.
The T-LSTM proposed by \citet{baytas17} improved on \citet{pham2016} by transforming time gaps into weights using a decay function and using them to adjust the hidden state. \citet{Li2019} improved on GRU-D \citep{che2016recurrent} by considering the observation rate in addition to the exponential decay on hidden state and weighted imputation. 
\citet{attain2019} showed that using a time-aware attention model outperforms \citet{Choi2016} and the T-LSTM on detection tasks.

Recurrent methods based on bidirectional RNNs typically achieve better performance than those based on single directional RNNs for both smoothing and classification tasks. \citet{Yoon2017} and \citet{Yoon18} proposed an MRNN based on bidirectional RNNs, which outperforms \citet{che2016recurrent}, \citet{futoma2017improved}, \citet{lipton2016directly} and \citet{Choi2015} on both smoothing and detection tasks. On the smoothing (imputation) task, it also outperforms standard methods such as  spline and cubic interpolation, MICE, Miss Forest, matrix completion and auto-encoder based approaches. \citet{brits2018} show that another bidirectional RNN-based approach for learning the missing values in vector-based representations   outperforms GRU-D and MRNN on both smoothing and classification tasks. 

GAN-based methods for imputing missing values in irregularly sampled time series data \citep{Luo2018,Luo2019} have been shown to achieve better classification performance than GRU-D. \citet{Luo2019} also outperforms \citet{Luo2018} and \citet{gain} on classification and smoothing tasks and \citet{brits2018} on classification tasks. The VAE-based approach of \citet{fortuin20} achieves better classification performance than \citet{Luo2018} and \citet{brits2018}.

ODE-based recurrent models typically achieve better classification, imputation, detection and prediction performance on irregularly sampled time series than discrete time RNN-based recurrent models. For example, Latent-ODE \citep{Rubanova2019} outperforms GRU-D, an RNN with exponential decay on hidden state, an RNN with imputation based on the weighted average of previous values and the empirical mean, and an RNN with values concatenated with time gaps and response indicators on both classification and detection tasks. Similarly, another related method based on the combination on ODE and RNN components \citep{DeBrouwer2019} achieves better forecasting performance than GRU-D and T-LSTM. Neural CDE \citep{kidger2020} achieves better performance than \citet{DeBrouwer2019}, GRU-D and ODE-RNN \citep{Rubanova2019} on whole time series classification tasks. 

Among interpolation based methods, IP-Nets introduced by \citet{Shukla2019} outperforms the multi-output Gaussian process regression model \citep{futoma2017improved}, GRU-D and RNN  baselines  based  on  mean  and  forward  imputation  schemes,  and  RNN models based on concatenating time gaps and missing indicators on classification and regression task. Attention-based interpolation \citep{shukla2020multitime} has been shown to achieve better performance than IP-Nets, GRU-D, Latent-ODE \citep{Rubanova2019}, Phased-LSTM \citep{Neil} and set-based approaches \citep{seft} on classification, detection and interpolation tasks. Finally, a dual-attention based time-aware method introduced by \citet{datagru2020} outperforms T-LSTM, GRU-D and IP-Nets on classification task. \citet{seft} showed that a set-based and transformer-based approach for modeling irregularly sampled time series achieve comparable performance to GRU-D and IP-Nets on classification tasks. 

In the next section, we provide a final distillation of the above evidence in terms of the strengths and weaknesses of different approaches and discuss promising directions for future research.



%% file: conclusions.tex
\section{Discussion and Directions for Future Work}

In this survey, we have discussed the relationship between the three fundamental representations of multivariate irregularly sampled time series (series-based, vector-based, and set-based), which motivate the use of different modeling primitives. We have introduced a categorization of approaches to modeling irregularly sampled time series data based on the fundamental modeling primitives used to accommodate irregular sampling. These modeling primitives include temporal discretization, interpolation, recurrence, attention and structural invariance. As we have shown, these modeling primitives underlie essentially all recent work on the modeling multivariate irregularly sampled time series. 

Importantly, different modeling primitives are imbued with different strengths and weaknesses as a result of the representations that they leverage. A key example of this is the case of recurrent methods that, by leveraging the vector-based representation, must explicitly tackle the problem of incomplete vector-valued irregularly sampled observations that the interpolation and set-based primitives avoid by leveraging representations that do not yield explicit missing data when multi-dimensional observations are unaligned.  

We have also defined a range of inference tasks that can be performed using irregularly sampled time series including detection, prediction, filtering, smoothing, interpolation and forecasting. As we have seen, there are also fundamental relationships between certain modeling primitives and the ability to successfully carry out particular tasks. Single-directional recurrent models and the interpolation/smoothing tasks are a prime example of this where the inherent structure of single-directional recurrent model typically results in maximally uncertain predictions immediately prior to the next observation. This is again an issue that is avoided by interpolation, attention and set-based methods that can inherently consider observations in the relative past and future of a given time point.

Finally, we have described a large number of specific models and methods categorized by the modeling primitives they build on. We believe this categorization is very useful as the models and methods based on particular modeling primitives inherit intrinsic strengths and weaknesses from the underlying modeling primitives. This includes strengths and weaknesses derived from the underlying time series representations and the relationships between modeling primitives and tasks. We summarize the identified strengths and weaknesses of each category of approaches below, as well as evidence regarding the relative predictive performance of approaches.

\begin{enumerate}
\item Discretization combined with imputation and any supervised model provides a simple, easy to implement, and modular baseline for solving whole time series classification and regression problems. The inclusion of response indicators as additional inputs has been found to help predictive performance under such an approach. However, the evidence indicates that such approaches are out-performed by methods based on other primitives. 

\item As described above, due to the inherent filtering nature of single-directional discrete recurrent models, basic discrete RNNs are more well suited for detection and prediction tasks than interpolation and smoothing problems. Bi-directional recurrent models typically out-perform single-directional models in terms of supervised tasks and can provide reasonable solutions to smoothing and interpolation problems. Using time stamps or time deltas as additional inputs appears to be helpful in overcoming the gap between representing a sequence and representing an irregularly sampled time series. However, RNNs must also deal with the problem of incomplete vector-valued inputs. Here, more sophisticated approaches for end-to-end learning of model components (such as decay mechanisms) that enable imputation appear to be helpful.

\item ODE-based recurrent approaches provide a more elegant solution to dealing with irregular sampling than discrete RNNs and can be applied to all tasks described in Section \ref{tasks}. ODE-based approaches have been shown to outperform discrete RNN-based models on several tasks including whole time series classification, interpolation, detection, and prediction. However, ODE-RNN models still have limitations related to their nature as intrinsic filtering methods as well as their inability to accommodate incomplete vector-valued observations. The related neural CDE-based model family appears to have a number of advantages over ODE-RNN models in terms of its ability to incorporate observations through time.

\item Interpolation-based models can also be applied to all tasks described in Section \ref{tasks}. Deterministic kernel smoothing-based methods have been shown to outperform RNN and ODE-based methods on several tasks including whole time series classification, interpolation, detection, and prediction. These models have also been shown to be substantially faster to train than Gaussian process regression and ODE-based models.   

\item Attention and structural invariance-based methods both break the sequential nature of recurrent approaches and have the potential for reductions in training time relative to recurrent models as a result. However, to date these models have been able to meet but not exceed the performance of ODE and interpolation-based methods.
\end{enumerate}

In terms of directions for future work, the attention and structural invariance modeling primitives have promising properties, but have been much less explored than other primitives including recurrence. The further development of attention and structural invariance-based approaches may lead to improved accuracy-speed trade-offs by leveraging the enhanced parallel computation that these primitives enable. Within the area of recurrent models, differential equation-based models for irregular time series have clearly advanced the state of the art over discrete RNNs. Neural CDEs appear to have interesting advantages over ODE-based models as noted above and are also an interesting area for further exploration. 

At the task level, it is apparent that recent research in the machine learning community has focused most heavily on supervised problems, followed by interpolation and smoothing. There has been significantly less attention on the forecasting task. Indeed, most of the methods reviewed in this paper have not been applied to forecasting tasks. The development of methods for learning to forecast accurately from irregularly sampled inputs thus appears to be a significantly more open problem than the development of methods for the other tasks described. 

In terms of performance evaluation, most of the methods considered here have  focused on criteria such as accuracy and mean squared error that do not account for uncertainty in model outputs. Indeed, models applied to classification tasks typically have not reported expected calibration error despite most models having the ability to produce distributions over classes \citep{guo2017calibration}. Further, most models applied to the interpolation and smoothing tasks do not produce probability distributions as outputs at all, apart from models based on Gaussian processes. Indeed, in the process of pursuing faster training times and improved flexibility in the multivariate setting, deterministic interpolation-based models have sacrificed the ability of Gaussian process-based models to translate input sparsity into uncertainty over predictive outputs. Restoring the ability to sensibly propagate uncertainty is a key challenge for current deterministic deep learning-based models. 

%% file: dataset.tex
\section{Data Sets}
\label{sec:datasets}
In this appendix, we describe some of the data sets that have been widely used in the machine learning community to conduct experiments with irregularly sampled time series and describe some of the commonly used tasks associated with each data set. We note that while it is possible to transform a regularly sampled time series into a sparse and irregularly sampled time series by randomly sampling a set of observations points, here we only describe data sets that naturally consist of sparse and irregularly sampled time series.

\subsection{MIMIC-III}
\label{sec:mimic}
The MIMIC-III data set \citep{johnson2016mimic} is a de-identified
data set collected at Beth Israel Deaconess Medical Center
from 2001 to 2012. It consists of approximately 58,000 hospital admission records.
This data set contains sparse and irregularly sampled physiological signals, medications, diagnostic codes, in-hospital mortality, length of stay, discharge summaries, progress notes, demographics information and more. Benchmark tasks on this data set include in-hospital mortality prediction, length of stay prediction and phenotype prediction. In-hospital mortality and length of stay prediction are whole time series classification and regression tasks while phenotype classification task is a multi-label classification task. The length of stay prediction task has also be framed as a classification task by discretizing the length of stay variable into a fixed number of buckets. MIMIC-III is available at \url{https://mimic.physionet.org/}.

\subsection{PhysioNet 2012}
\label{sec:physionet}
The PhysioNet Challenge 2012 data set \citep{physionet} consists of multivariate time series data with $37$ variables extracted from intensive care unit (ICU) records. Each record contains sparse and irregularly spaced measurements from the first $48$ hours after admission to ICU. The data set includes $4000$ labeled instances and $4000$ unlabeled instances.  Benchmark tasks on this data set are in-hospital mortality prediction (whole time series classification) and interpolation. All $8000$ instances can be used for interpolation experiments while only $4000$ labeled instances are available for classification experiments. The PhysioNet dataset is available at \url{https://physionet.org/content/challenge-2012/}.

\subsection{Human Activity}
\label{sec:human_activity}
The Localization Data for Human Activity data set \citep{humanactivity} consists of 3D positions of the waist, chest and ankles collected from five individuals performing several activities including walking, sitting, lying, standing, etc. 
Following the data preprocessing steps of \citet{Rubanova2019}, a data set of $6,554$ sequences with $12$ channels and $50$ time points can be constructed. Labels are provided for each observation time point and denote the type of activity that the person is performing, such as walking, sitting, lying, etc. The data set consists of 11 classes. A per-time point classification task (detection task) is the benchmark task on this data set. The data set is available at \url{https://archive.ics.uci.edu/ml/datasets/Localization+Data+for+Person+Activity}.

\subsection{eICU Collaborative Research Data Set}
The eICU Collaborative Database \citep{pollard2018} is a database relating to patients who were treated as part of the Philips eICU program across intensive care units in the United States. 
eICU consists of medical records from 200,859 patients collected from 208 critical care units in the United States between 2014 and 2015. 
The database is deidentified, and includes sparse and irregularly sampled vital sign measurements, care plan documentation, severity of illness measures, diagnosis information, treatment information, and more.
The data set is available at \url{https://eicu-crd.mit.edu/}.

\subsection{PhysioNet 2019}
PhysioNet 2019 \citep{reyna2019} is a publicly available data set collected from two hospital systems. It contains 40,336 ICU patient admission records and 2,359 records of diagnosed sepsis cases. The data set consists of a set of multivariate time series that contains 40 variables including 8 vital signs, 26 laboratory values and 6 demographic variables. Each timestamp is labeled with a binary variable indicating whether the onset of sepsis has occurred. The benchmark task on this data set is the early detection of sepsis using physiological data.

%% file: appendix.tex
\section{Source Code}
In this appendix, we provide links to implementation source code for many of the methods discussed in this paper. This information is presented in Table  \ref{tab:my_label}.

\label{sec:code}
\begin{table}[h]
\caption{Source code for different approaches}
    \label{tab:my_label}
    \centering
    \begin{tabular}{l|P{0.7\linewidth}}
    \hline
        {\bf Reference} & {\bf Source Code} \\
         \hline
\citet{baytas17}	& \url{https://github.com/illidanlab/T-LSTM} \\
\citet{brits2018}	& \url{https://github.com/caow13/BRITS} \\
\citet{che2016recurrent}	& \url{https://github.com/Han-JD/GRU-D} \\
\citet{neural_ode2018}	& \url{https://github.com/rtqichen/torchdiffeq} \\
\citet{Choi2015}	& \url{https://github.com/mp2893/doctorai} \\
\citet{Choi2016}	& \url{https://github.com/mp2893/retain} \\
\citet{DeBrouwer2019}	& \url{https://github.com/edebrouwer/gru_ode_bayes} \\
\citet{fortuin20}	& \url{https://github.com/ratschlab/GP-VAE} \\
\citet{benchmark}	& \url{https://github.com/YerevaNN/mimic3-benchmarks} \\
\citet{seft}	& \url{https://github.com/BorgwardtLab/Set_Functions_for_Time_Series} \\
\citet{Kim2018}	& \url{	https://github.com/ykim32/TBM-missing-data-handling	} \\
\citet{li2016scalable}	& \url{https://github.com/steveli/gp-adapter} \\
\citet{Li2020} & \url{https://github.com/steveli/partial-encoder-decoder} \\
\citet{Luo2019}	& \url{https://github.com/Luoyonghong/E2EGAN} \\
\citet{Neil}	& \url{https://github.com/dannyneil/public_plstm} \\
\citet{pham2016} & \url{https://github.com/trangptm/DeepCare}\\
\citet{Rubanova2019}	& \url{https://github.com/YuliaRubanova/latent_ode} \\
\citet{Shukla2019}	& \url{https://github.com/mlds-lab/interp-net} \\
\citet{attendanddiagnose2018}	& \url{	https://github.com/khirotaka/SAnD	} \\
\citet{Yoon2017}	& \url{	https://github.com/jsyoon0823/MRNN	} \\
\citet{selfatttimeemb2019}	& \url{	https://github.com/StatsDLMathsRecomSys/Self-attention-with-Functional-Time-Representation-Learning	} \\

\hline
    \end{tabular}
\end{table}